\documentclass[manuscript,nonacm]{acmart}

\AtBeginDocument{%
  }

\usepackage{graphicx}

\begin{document}

\title{Overcoming Data Scarcity in Generative Language Modelling for Low-Resource Languages: A Systematic Review}

\author{Josh McGiff}
\email{josh.mcgiff@ul.ie}
\author{Nikola S. Nikolov}
\affiliation{%
  \institution{Department of Computer Science and Information Systems, University of Limerick}
  \city{Limerick}
  \country{Ireland}
}


\begin{abstract}

Generative language modelling has surged in popularity with the emergence of services such as ChatGPT and Google Gemini. While these models have demonstrated transformative potential in productivity and communication, they overwhelmingly cater to high-resource languages like English. This has amplified concerns over linguistic inequality in natural language processing (NLP).
This paper presents the first systematic review focused specifically on strategies to address data scarcity in generative language modelling for low-resource languages (LRL). Drawing from 54 studies, we identify, categorise and evaluate technical approaches, including monolingual data augmentation, back-translation, multilingual training and prompt engineering, across generative tasks. We also analyse trends in architecture choices, language family representation, and evaluation methods. Our findings highlight a strong reliance on transformer-based models, a concentration on a small subset of LRLs, and a lack of consistent evaluation across studies. We conclude with recommendations for extending these methods to a wider range of LRLs and outline open challenges in building equitable generative language systems. Ultimately, this review aims to support researchers and developers in building inclusive AI tools for underrepresented languages, a necessary step toward empowering LRL speakers and the preservation of linguistic diversity in a world increasingly shaped by large-scale language technologies.

\end{abstract}




\maketitle

\section{Introduction}

Language modelling is a central task in the field of natural language processing (NLP) that aims to capture the features of a language by extracting the statistical and contextual relationships within a language \cite{wang2023pre}. This information can subsequently be used to inform next-word predictions for a given sequence of previous terms \cite{klosowski2018deep}. It involves using a model to extract the characteristics of a language from a large corpus of textual data that can consequently be used to process, understand and generate text \cite{thirunavukarasu2023large}. Language models evolved from statistical methods like Markov processes \cite{almutiri2022markov} and n-gram models \cite{roark2007discriminative} to recurrent neural network (RNN) approaches, such as Long Short-Term Memory (LSTM) RNNs \cite{van2020review}. However, the current rise in popularity of language models and language model-powered services such as ChatGPT \cite{lian2024public,ng2024powerful} can be attributed to the breakthrough that introduced the transformer \cite{vaswani2017attention} as an architecture for modelling languages. This advancement enabled the development of pre-trained models such as the transformative Generative Pre-trained Transformer (GPT) \cite{radford2018improving} and Bidirectional Encoder Representations from Transformers (BERT) \cite{devlin2018bert}, with the transformer architecture enhancing NLP tasks such as machine translation \cite{raganato2018analysis}, classification \cite{mcgiff2024bridging} and named entity recognition \cite{arkhipov2019tuning}. 

Generative language modelling is a subset of language modelling that focuses on building models that can produce new sentences for a set of given semantics \cite{su-etal-2019-dual}. Generative language modelling is associated with the topic of natural language generation (NLG) as natural language text can be produced for a variety of communicative goals \cite{dong2022survey}. This differs from natural language understanding (NLU) whereby the morphology, syntax and semantics are processed to comprehend the meaning of text, identifying intent or extracting information \cite{xu2020curriculum}. Generative language models are the driving force behind popular services such as ChatGPT \cite{lian2024public,ng2024powerful}, as they process a prompt and produce natural language text that is aligned with the context of the input.  These models are flexible in their ability to represent several languages at once, with studies successfully modelling both individual languages \cite{de2020geppetto,muller2022cedille,simoulin2021modele} and groups of languages \cite{liu2020multilingual,xue2021mt5}. 

NLP research has been dominated by a minority of the world's languages \cite{nigatu2024,joshi-etal-2020-state,bird-2022-local}. Nigatu \textit{et al.} describe the 'low-resource' and 'high-resource' categories that the NLP community has adopted to classify the research interest for a given language \cite{nigatu2024}. Their study echoes the reality that low-resource languages (LRL) are constantly trailing the NLP developments and breakthroughs experienced by high-resource languages. Although "low-resource" can be interpreted as less studied and less computerised \cite{magueresse2020low}, it can also be used to describe a language with scarce resources \cite{ogueji2021small}. Some evidence suggests that the availability of NLP researchers with fluency in a language correlates with the availability of datasets \cite{yu-etal-2022-beyond}. This connects the theoretical challenge of sourcing data for language modelling with the challenge of ensuring that LRL speakers are not left behind in technological advancements. Although these LLM-based services can improve worldwide communication and information access, their focus on building and optimising their applications for English could hinder linguistic diversity \cite{zhang2023ethical,kirk2023personalisation}. In terms of language preservation, some research has connected generative language models with being capable of revitalising endangered languages \cite{nanduri2023revitalizing} and conserving languages that are less predominantly modelled by the major LLM providers \cite{hutson2024preserving,lai-etal-2023-chatgpt}. Therefore, developing a generative language model that accurately captures the linguistic features of an LRL could serve as a digital preservation tool, mitigating the risk of its extinction. The need to support LRLs with modern technological advancements has motivated some research groups that are working towards improving LRL representation in NLP and NLG research \cite{singh2024aya, nllb2024scaling,le2023bloom}. 

A large volume of data is required to build a robust generative language model \cite{kaplan2020scaling, NEURIPS2023_9d89448b,hoffmanntraining,radford2019language}. This conflicts with the limited textual data that is available to LRLs. Many studies aim to tackle the data disparity, with one main branch of these approaches focusing on integrating a data augmentation process into the model training pipeline \cite{shorten2021text}. Data augmentation is a process that generates synthetic data from an existing dataset \cite{shorten2019survey}. The concept of data augmentation is commonly used in the field of computer vision to train more robust models \cite{taylor2018improving, mikolajczyk2018data, khosla2020enhancing}. Image datasets are often enlarged by performing image transformations such as rotations or brightness modifications to create additional training images \cite{mumuni2022data}. While it has been argued that NLP has not adopted data augmentation to the same degree as computer vision, a wide range of studies have adopted a variety of processes to bridge the gap in available training data for LRL modelling \cite{shorten2021text}. 
Although prior surveys have explored data augmentation in NLP broadly, no systematic review to date has focused on how these strategies are applied in the development of generative models for LRLs. 

To address this gap, this paper presents the first systematic review focused specifically on techniques for overcoming data scarcity in the context of generative language modelling for low-resource languages. By analysing 54 studies, this review provides a structured overview of the technical approaches used, the language families studied, and the model architectures employed. It also highlights underrepresented languages, evaluates the effectiveness of key methods, and identifies research gaps in the development of generative models for LRLs. Therefore, the objectives of this systematic review are to: 
\begin{itemize}
    \item Identify languages and language families in particular that are underrepresented in this area of LRL modelling.
    \item Determine the methods that have been used to mitigate data scarcity, including but not limited to data augmentation in the development of generative language models for LRLs.
    \item Analyse the impact of methods for overcoming data scarcity on the performance of various language modelling tasks.
    \item Identify existing gaps in the literature and share insights relating to future research directions in generative LRL modelling.
\end{itemize}

\section{Review Methodology}
A systematic review is described by Page \textit{et al.} as being a review that uses clear, methodical techniques to aggregate and summarise the results of studies aimed at answering a clearly defined research question  \cite{page2021prisma}. It is an informative process that can yield in-depth exploration and determine the current state of existing knowledge in a field. The PRISMA 2020 statement indicates that a systematic review can combine information relating to future research opportunities, answer research questions that span different individual studies, identify shortcomings in research and explore existing primary research results \cite{page2021prisma}. The PRISMA review methodology can be divided into four steps for this review: (1) Defining the research objectives/questions and eligibility criteria; (2) Adopting a search strategy; (3) Performing the study selection; and (4) Data extraction. 
    \subsection{Defining the Research Questions and Eligibility Criteria}
    This systematic review focuses on the main research objective of creating generative language models for LRLs using methods to mitigate the challenges posed by limited data. However, we have identified several research sub-questions that will help to inform this overarching research objective.

    Defining explicit criteria for including or excluding evidence is necessary for enabling readers to both understand the scope of the review and to verify eligibility decisions \cite{page2021prisma}. We apply the Population, Intervention, Comparison, and Outcome (PICO) framework to our research objective to define the eligibility criteria for evidence in this systematic review \cite{stone2002popping}. The PICO framework consists of four key elements: Population (P), which defines the target group under study; Intervention (I), which refers to the treatment, technique, or approach being evaluated; Comparison (C), which contrasts the intervention against an alternative or baseline; and Outcome (O), which represents the measurable effects or improvements resulting from the intervention. The PICO framework is employed to formulate a well-defined research question, which serves as the foundation for constructing a systematic database search string in accordance with the PRISMA methodology. LRLs are considered to be the population of the framework. On the other hand, the techniques for overcoming data scarcity are considered to be the intervention element of the framework. The comparison section of the framework is linked to comparing the performance of generative language models, both when using data augmentation or other technical methods to address data disparity for LRLs, and when not employing these methods. The outcome element of the framework relates to quantitative (eg: sacreBLEU, COMET etc) and qualitative (eg: human feedback) improvement in language generation. The PICO framework produces the following revised research question: In the context of LRLs (Population), how do techniques designed to overcome data scarcity (Intervention) compare to traditional NLG methods (Comparison) in improving the performance of generative language models (Outcome)?
    
    Once the revised research question was formulated using the PICO framework, we devised a set of inclusion and exclusion criteria to filter out the most irrelevant studies from our search. We included the papers that satisfied at least one of the following inclusion criteria:
\begin{enumerate}
    \renewcommand{\labelenumi}{IC\arabic{enumi}.}
    \item The paper focuses on building generative language models for LRLs.
    \item The paper uses a technical method for overcoming data scarcity.
    \item The paper reports their methods or reviews existing methods that tackle generative language modelling for LRLs.
\end{enumerate}

Additionally, we excluded the papers that satisfied at least one of the following exclusion criteria: 
\begin{enumerate}
    \renewcommand{\labelenumi}{EC\arabic{enumi}.}
    \item Papers not written in English. 
    \item We excluded all papers not published in journals, conference proceedings, workshops, symposia, or as arXiv preprints, and we also excluded posters and abstracts.
    \item Papers that create language models for NLP-specific tasks as opposed to NLG tasks.
    \item Papers published before 2015, or preprints released before 2021.
    \item Papers that do not produce results or survey existing literature relating to creating NLG models for LRLs or data augmentation techniques.
\end{enumerate}

\subsection{Adopting a Search Strategy}
In terms of the search strategy for this systematic review, we initially compiled a list of information sources for the search itself. The strategy involved searching SpringerLink, ACL Anthology, ACM, IEEE Xplore, Scoupus, Wiley Interscience and Google Scholar. These platforms were chosen for their expansive digital repositories. Specific sources, such as the ACL Anthology, were included due to their relevance to the computational linguistics focus of the research objective. Table~\ref{tab:search_string} details the search strings that were used for the various digital repositories. It is worth flagging that some of the platforms differ in terms of the syntax required for building search strings. Therefore, we report the equivalent strings for each platform. The searches match elements in the search string with metadata associated with each study such as the title, keywords and abstract. Most search engines use a combination of logical OR and logical AND operations to combine different terms and to impose constraints on groups of terms respectively. Search string keywords were chosen to match the main terms in the PICO research objective and sub-questions. Synonyms for terms such as "low resource language" and "language model" were sourced and included in the search string to further ensure the retrieval of relevant studies.       

\begin{table}[htbp]
    \centering
    \renewcommand{\arraystretch}{1.3} 
    \setlength{\tabcolsep}{10pt} 
    \caption{Search string composition as part of the search strategy.}
    \label{tab:search_string}
    \begin{tabular}{|p{3cm}|p{10cm}|}
        \hline
        \textbf{Scope} & \textbf{String} \\
        \hline
        Generative Aspect & "generative" \\
        \hline
        Model Type & "language model" OR "text model" \\
        \hline
        Language Type & "low resource language" OR "minority language" OR "endangered language" \\
        \hline
        Exclusion & "classification" \\
        \hline
    \end{tabular}
\end{table}

\subsection{Performing the Study Selection}
The study selection phase of this research was initially carried out using ASReview, an open-source machine learning-powered tool for performing efficient systematic reviews \cite{van2021open}. The study was subsequently completed with a manual review of the remaining papers. ASReview assists researchers by predicting the most relevant studies and ranking them accordingly. We found that this accelerated the study selection process. The ASReview element of the selection process involved loading a dataset of bibliographic records from the previously listed information sources in RIS format. Once the dataset of papers was successfully imported, we chose TF-IDF for feature extraction, Naive Bayes as the classifier, ASReview's maximum query strategy and dynamic resampling as the balance strategy. The systematic review tool begins the screening process by displaying a random sample of studies to initialise the active learning element. Studies can be labelled as being relevant or irrelevant with these classifications being used to predict the next most relevant paper. The papers that are most likely to be relevant are prioritised for review. It is necessary to report that a single reviewer worked independently on each stage of the screening process. This process was used for the first pass through the original dataset of papers collected from the previously defined information sources. The first pass removed all papers that satisfied EC1, EC3 and EC4. The second pass involved screening papers that satisfied EC2 or EC5. This involved manually reading the remaining papers and adding relevant papers found via reference tracking to the list of included studies. 

\subsection{Data Extraction}
The data collection process involved extracting information relating to qualitative and quantitative features of the studies that align with the objective and research questions of this systematic review. The attributes extracted from the studies are formulated into data items represented via data rows in a spreadsheet. In terms of the quantitative information, the data items included: the title of the paper; whether it is a journal, conference, workshop, symposium or preprint paper; the publication name and a list of results for quality assessment questions. The quality assessment questions were formulated following PRISMA guidelines and existing review papers \cite{peixoto2021immersive,zapata2015empirical,sardi2017systematic}. The purpose of the quality assessment questions is to help gauge the relevance and value of the studies in relation to the research objective of the systematic review. We report the following quality assessment questions:

\begin{enumerate}
\renewcommand{\labelenumi}{QA\arabic{enumi}.}
\item Does the paper discuss building language models for LRLs? (+1) Yes/ (+0) No
\item Does the paper present empirical results?  (+1) Yes/ (+0) No
\item Does the paper discuss a process for overcoming data scarcity? (+1) Yes/ (+0) No
\item Has the paper presented and compared the results of using data augmentation? (+1) Yes/ (+0) No
\item Was the paper published in a relevant journal or conference proceedings? Given that there are different metrics for categorising the relevance of journal and conference publications, we consulted the Journal Citation Reports 2024 and CORE 2023 to determine a suitable score for each study. For conferences, studies were scored based on the CORE 2023 ranking: +1.5 for CORE A, +1 for CORE B, +0.5 for CORE C, and +0 for unranked conferences. For journals, scores were assigned using the Journal Citation Reports 2024: +2 for Q1, +1.5 for Q2, +1 for Q3/Q4, and +0 for unranked journals. Other publications received a score of +0. 
\end{enumerate}

     In terms of the qualitative information, the data items included attributes representing each of the research sub-questions that were formulated to better inform the overall research objective of this systematic review. We report the following qualitative research sub-questions for data extraction:

     \begin{enumerate}
    \renewcommand{\labelenumi}{RQ\arabic{enumi}:}
    \item What low-resource languages have already been modelled by other studies? This should consist of a list of languages that the study models. While some studies could model a single language, others could model a range of languages. 
    \item What technical methods have been used to overcome data scarcity in building generative language models? Answers for this question should represent categories relating to specific technical methods. 
    \item What text data augmentation techniques have been used to build generative language models for low-resource languages? This section should include all the text data augmentation techniques that produce new data from existing textual data.
    \item Which sources are publishing in the area of low-resource language modelling? This attribute should consist of the name of the publisher of the study. 
    \item What architectures are being used to produce low-resource language models? This research question should produce a list of the model architectures that are used by the study to model LRLs.
    \item How effective have methods for overcoming data scarcity been at improving model performance for different NLG tasks? This attribute should include various metrics that were produced to evaluate the NLG models.
    \item How scarce is the data in studies tackling low-resource language modelling? This should consist of a quantity that describes the size of the dataset being used for training generative language models for LRLs. 
    \item How are the models in these studies being evaluated? This research question should explore the evaluation methods that are being used to measure the performance of generative language models for LRLs.
    \item What is the NLG task? This section should include the categories of NLG tasks that the study tackles for LRLs. 
\end{enumerate}

\section{Results}
This section is structured into two parts: the first presents the outcomes of the study selection process and the second details the findings related to the extraction of research question-specific characteristics from the included studies. 

    \subsection{Study Selection}
\begin{figure}[htbp]
        \centering
        \includegraphics[width=1\textwidth]{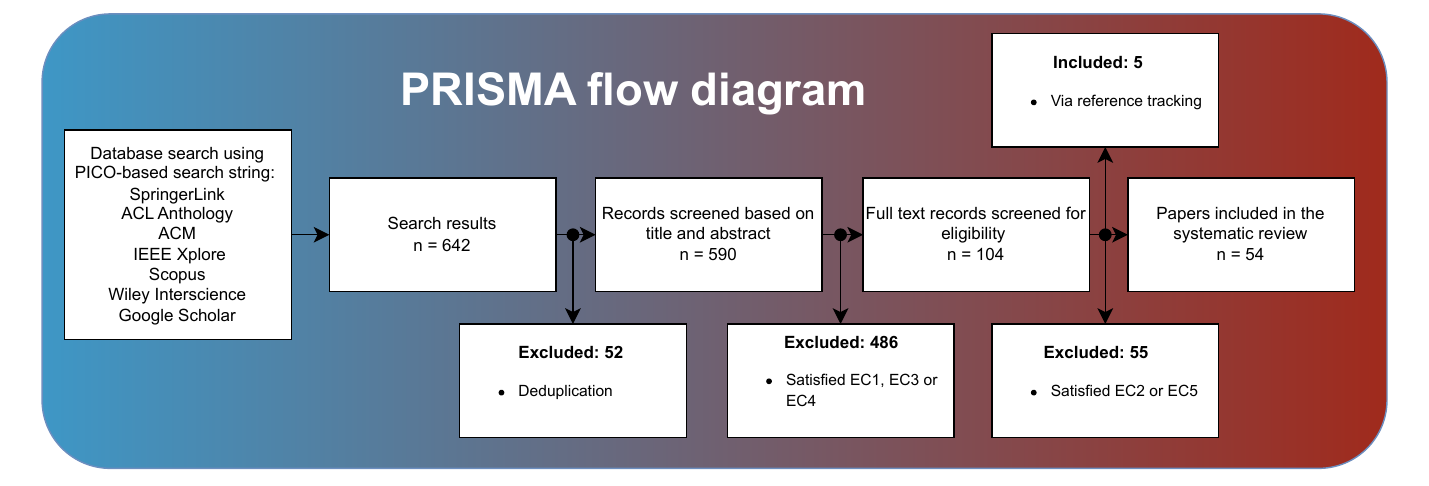}
        \caption{This PRISMA flow diagram describes the process of using the PICO framework-based search string to find and filter relevant studies. After applying each exclusion criteria and including additional papers via reference tracking, 54 papers were included in the systematic review.}
        \label{fig:prisma}
    \end{figure}

    The study selection section of this systematic review was split into three phases. A set of 642 papers were found by searching the previously mentioned information sources. Performing deduplication on the dataset removed 52 papers from the set and left 590 studies for screening. The first pass using ASReview for the screening process resulted in the rejection of 486 papers based on their title, keywords, abstract and whether they satisfy EC1, EC3, or EC4. The first pass retained 104 of the studies. The second pass screened for papers that satisfied EC2 or EC5. This second pass rejected 55 papers, retaining 49 papers in total. However, five papers were identified via reference tracking and consequently brought the final total to 54 papers for review. Figure~\ref{fig:prisma} represents the PRISMA flow diagram that reflects the screening process used to select the studies to include in this systematic review. 
    
\begin{figure}[htbp]
    \centering
    \includegraphics[width=0.8 \textwidth]{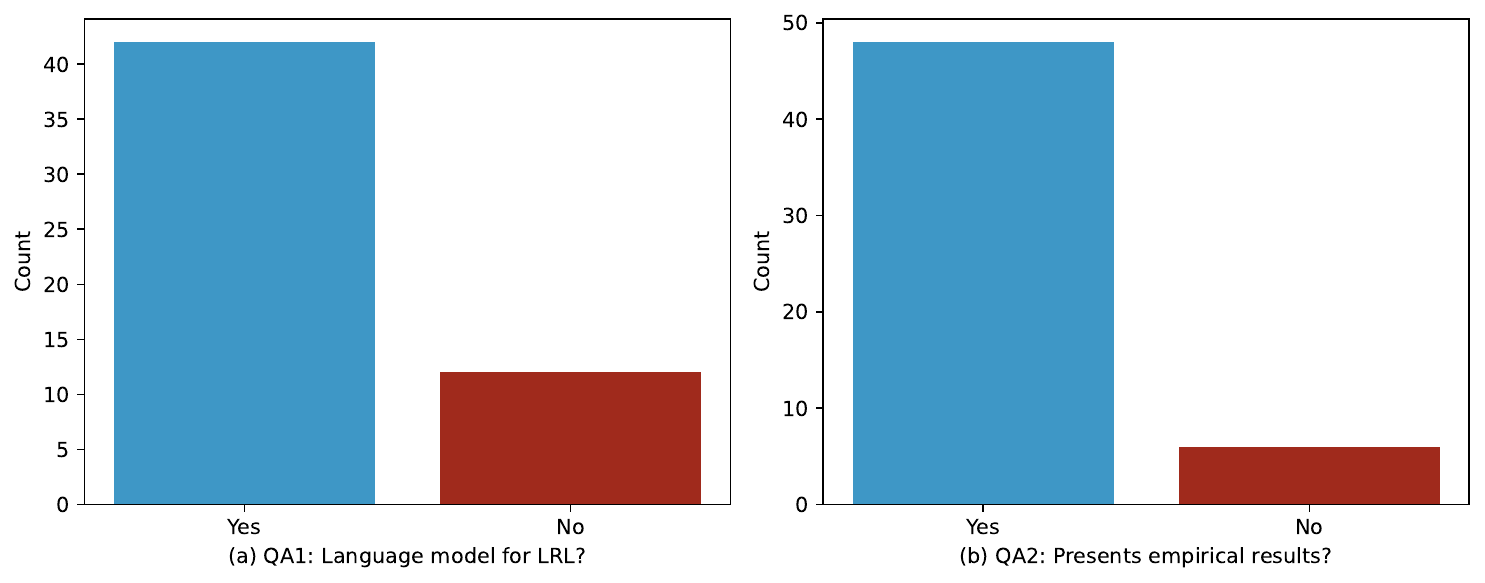}
\caption{Distribution of papers based on whether they tackle building a language model for a low-resource language (QA1), and whether they present empirical results (QA2).}
    \label{fig:qa1}
\end{figure}

\begin{figure}[htbp]
    \centering
    \includegraphics[width=0.8\textwidth]{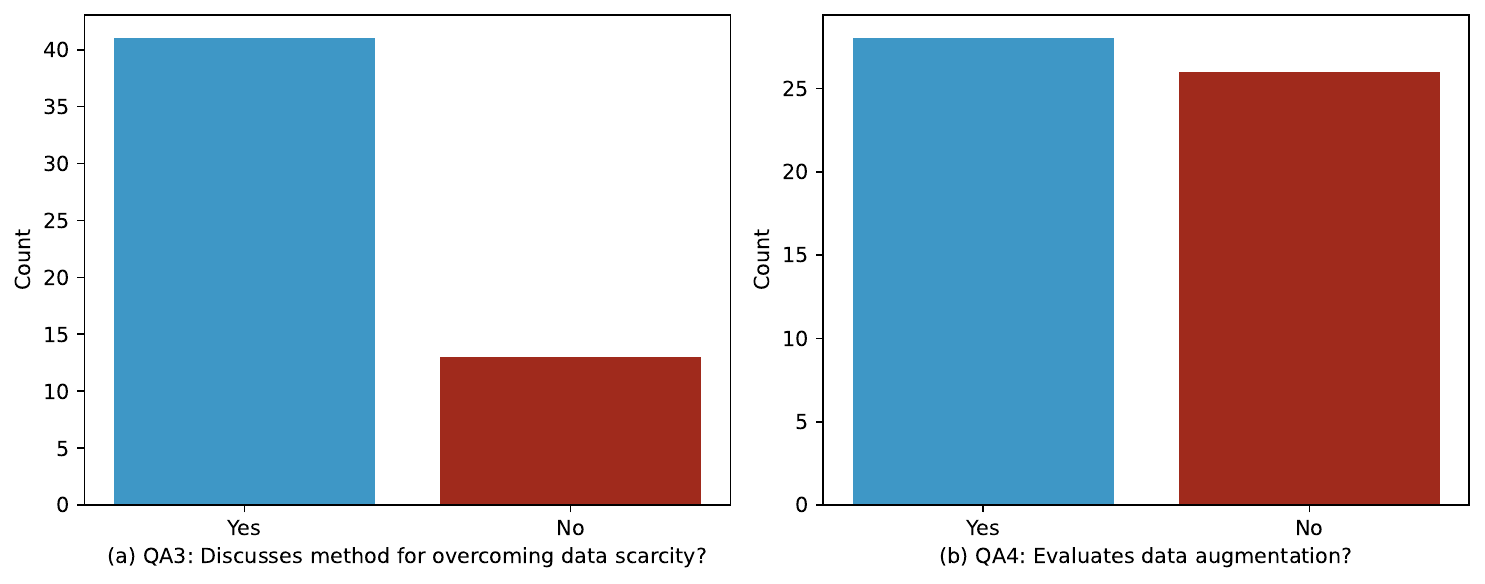}
\caption{Distribution of papers based on whether they discuss methods for addressing data scarcity (QA3), and whether they present and compare results of such methods (QA4).}
    \label{fig:qa2}
\end{figure}

\begin{figure}[htbp]
    \centering
    \includegraphics[width=0.8\textwidth]{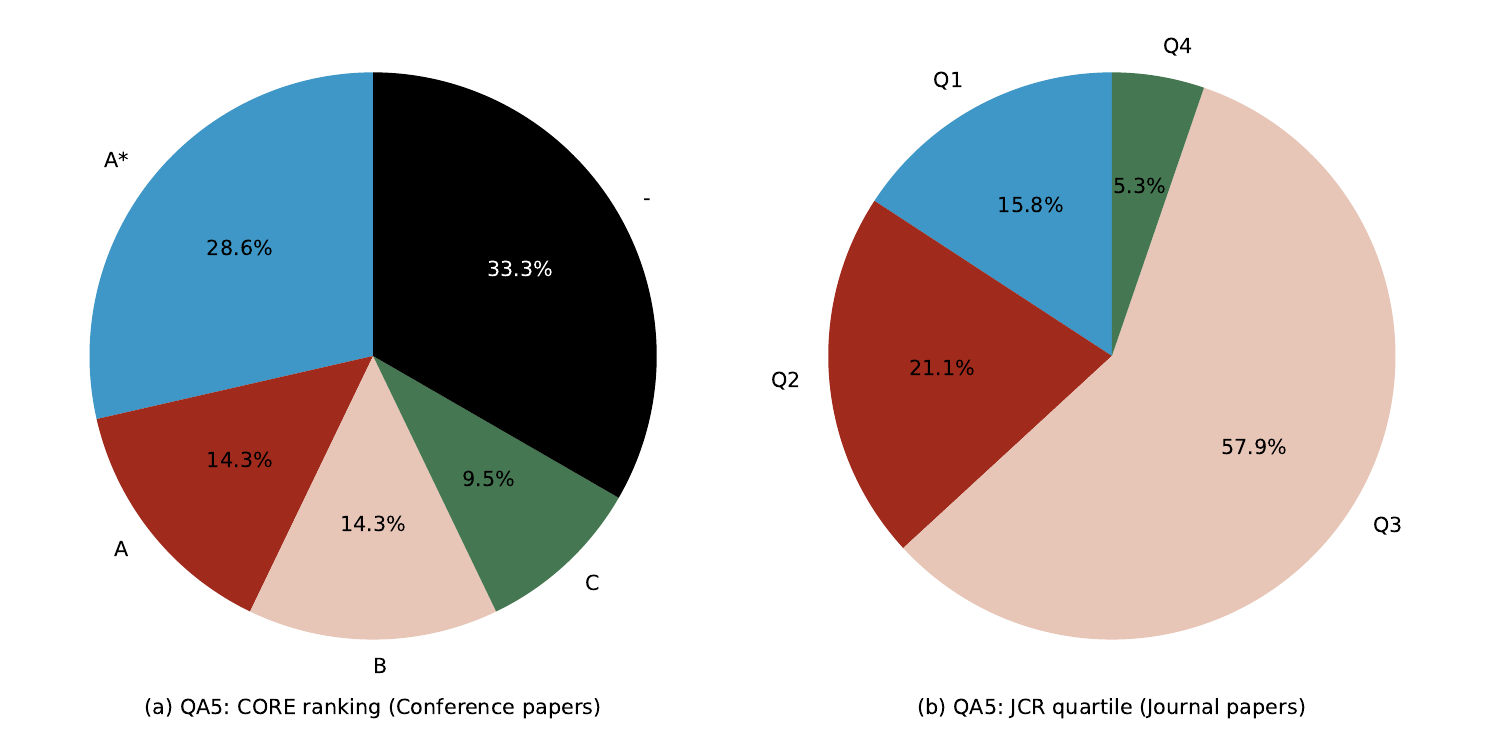}
    \caption{CORE ranking distribution for conference papers and  Journal Impact Factor 2024 distribution for journal papers (QA5).}
    \label{fig:core}
\end{figure}

    The 54 selected studies were subsequently evaluated using the predefined quality assessment criteria. The results of applying quality assessment are illustrated in Figures~\ref{fig:qa1}–\ref{fig:core}. Table~\ref{tab:qa} reports the quality assessment score assigned to each paper. The majority of papers in this review (77.78\%) tackle building language models for a LRL. 88.89\% of the literature present empirical results. Additionally, 75.93\% of studies discuss the use of a process for overcoming data scarcity. Over half of the studies (51.85\%) present and compare methods aimed at addressing data scarcity. Among the conference papers, 28.6\% were published in A* venues, 14.3\% in A-ranked and B-ranked venues respectively, 9.5\% in C-ranked venues, and 33.3\% were unranked, based on CORE conference rankings. Among the journal papers, 15.8\% were classified as Q1, 21.1\% as Q2, 57.9\% as Q3, and 5.3\% as Q4, based on the 2024 Journal Impact Factor quartile rankings.

\begin{table*}[t]
    \caption{Overview of data extracted for each quality assessment criterion: QA1–QA5. QA1: Does the paper discuss building language models for low-resource languages (LRLs)? QA2: Present empirical results? QA3: Discuss a method for overcoming data scarcity? QA4: Present and compare data augmentation results? QA1–QA4 are each scored as +1 for Yes, +0 for No. QA5 assesses publication venue quality: conferences are scored using the CORE 2023 ranking (+1.5 for CORE A, +1 for B, +0.5 for C, +0 for unranked); journals follow Journal Citation Reports (JCR) 2024 (+2 for Q1, +1.5 for Q2, +1 for Q3/Q4, +0 for unranked).}
        \label{tab:qa}
        \centering
        {\scriptsize  

        \begin{tabular}{p{0.5cm} p{1.5cm} p{1.5cm} c c c c c r} 
            \toprule
        Paper & Publication Year & Publication Type & QA1 & QA2 & QA3 & QA4 & QA5 & Total Score \\
        \midrule
    \cite{li_improving_2024} & 2024 & conference & y & y & y & y & A & 5.50\\
    \cite{noauthor_qugan_nodate} & 2019 & journal & y & y & y & y & Q2 & 5.50\\
    \cite{zhang_chren_2020} & 2020 & conference & y & y & y & y & A* & 5.50\\
    \cite{usui_translation_2023} & 2023 & conference & y & y & y & y & A & 5.50\\
    \cite{sorokin_ask_2022} & 2022 & conference & y & y & y & y & A & 5.50\\
    \cite{gao_soft_2019} & 2019 & conference & y & y & y & y & A* & 5.50\\
    \cite{bhowmick_improving_2023} & 2023 & journal & y & y & y & y & Q3 & 5.00\\
    \cite{otegi_conversational_2020} & 2020 & conference & y & y & y & y & B & 5.00\\
    \cite{marie_synthesizing_2020} & 2020 & journal & n & y & y & y & Q1 & 5.00\\
    \cite{ahmadnia_augmenting_2019} & 2019 & journal & y & y & y & y & Q3 & 5.00\\
    \cite{yirmibesoglu_morphologically_2023} & 2023 & journal & y & y & y & y & Q3 & 5.00\\
    \cite{maimaiti_improving_2021} & 2021 & journal & y & y & y & y & Q3 & 5.00\\
    \cite{cahyawijaya_indonlg_2021} & 2021 & conference & y & y & y & n & A* & 4.50\\
    \cite{guo_automatically_2021} & 2021 & conference & y & y & y & y  & A* & 4.50\\
    \cite{wongso_many--many_2023} & 2023 & journal & y & y & y & n & Q2 & 4.50\\
    \cite{dandapat_iterative_2018} & 2018 & conference & y & y & y & y & C & 4.50\\
    \cite{wang_switchout_2018} & 2018 & conference & n & y & y & y & A* & 4.50\\
    \cite{noauthor_paraphrase_nodate} & 2020 & conference & n & y & y & y & A* & 4.50\\
    \cite{pham_meta_2021} & 2021 & preprint & y & y & y & y & - & 4.00\\
    \cite{niyogi_paramanu_2024} & 2024 & preprint & y & y & y & y & - & 4.00\\
    \cite{downey_targeted_2024} & 2024 & preprint & y & y & y & y & - & 4.00\\
    \cite{hong_cantonmt_2024} & 2024 & preprint & y & y & y & y & - & 4.00\\
    \cite{chang_when_2023} & 2023 & preprint & y & y & y & y & - & 4.00\\
    \cite{shafayat_benqa_2024} & 2024 & preprint & y & y & y & y & - & 4.00\\
    \cite{tanwar_translating_2020} & 2020 & journal & y & y & y & n & Q3 & 4.00\\
    \cite{gerz_language_2018} & 2018 & journal & y & y & n & n & Q1 & 4.00\\
    \cite{berckmann_low-resource_2020} & 2020 & conference & y & y & y & y & - & 4.00\\
    \cite{liu_low-resource_2022} & 2022 & workshop & y & y & y & y & - & 4.00\\
    \cite{pham_van_improving_2022} & 2022 & symposium & y & y & y & y & - & 4.00\\
    \cite{nissanka_exploring_2020} & 2020 & conference & y & y & y & y & - & 4.00\\
    \cite{agarwal_zero-shot_2022} & 2022 & workshop & y & y & y & y & - & 4.00\\
    \cite{scalvini_evaluating_2024} & 2024 & conference & y & y & y & n & B & 4.00\\
    \cite{guo_teaching_2024} & 2024 & conference & y & y & y & n & B & 4.00\\
    \cite{noauthor_frontiers_nodate} & 2023 & journal & y & y & n & n & Q2 & 3.50\\
    \cite{bendel_llegra_2024} & 2024 & journal & y & y & n & n & Q2 & 3.50\\
    \cite{wu_study_2023} & 2023 & conference & n & y & y & y & - & 3.00\\
    \cite{babaali_breaking_2024} & 2024 & journal & y & y & n & n & Q4 & 3.00\\
    \cite{li_multi-tasking_2022} & 2022 & journal & y & y & n & n & Q3 & 3.00\\
    \cite{wang_survey_2022} & 2022 & journal & n & n & y & n & Q1 & 3.00\\
    \cite{mi_multi-granularity_2024} & 2024 & journal & y & y & n & n & Q3 & 3.00\\
    \cite{noauthor_efficient_nodate} & 2020 & journal & n & y & y & n & Q3 & 3.00\\
    \cite{noauthor_english-arabic_nodate} & 2023 & conference & n & y & y & n & C & 2.50\\
    \cite{mao_tuning_2024} & 2024 & workshop & y & y & n & n & - & 2.00\\
    \cite{toraman_llamaturk_2024} & 2024 & preprint & y & y & n & n & - & 2.00\\
    \cite{zhang_teaching_2024} & 2024 & preprint & y & y & n & n & - & 2.00\\
    \cite{acikgoz_bridging_2024} & 2024 & preprint & y & y & n & n & - & 2.00\\
    \cite{haque_recent_2021} & 2021 & journal & n & n & y & n & Q3 & 2.00\\
    \cite{shi_low-resource_2022} & 2022 & journal & n & n & y & n & Q3 & 2.00\\
    \cite{zhang_neural_2024} & 2024 & journal & n & n & y & n & Q3 & 2.00\\
    \cite{noauthor_banglagpt_nodate} & 2023 & conference & y & y & n & n & - & 2.00\\
    \cite{noauthor_bidirectional_nodate} & 2023 & conference & y & y & n & n & - & 2.00\\
    \cite{noauthor_qasina_nodate} & 2023 & conference & n & y & y & n & - & 2.00\\
    \cite{noauthor_generative_nodate} & 2023 & conference & y & n & n & n & - & 1.00\\
    \cite{wang2021survey} & 2021 & preprint & n & n & y & n & - & 1.00\\
    \bottomrule
        \end{tabular}
        }
    \end{table*}

    \subsection{RQ1: What Low-Resource Languages Have Already Been Modelled by Other Studies?}
\begin{figure}[htbp]
        \centering
        \includegraphics[width=1 \textwidth]{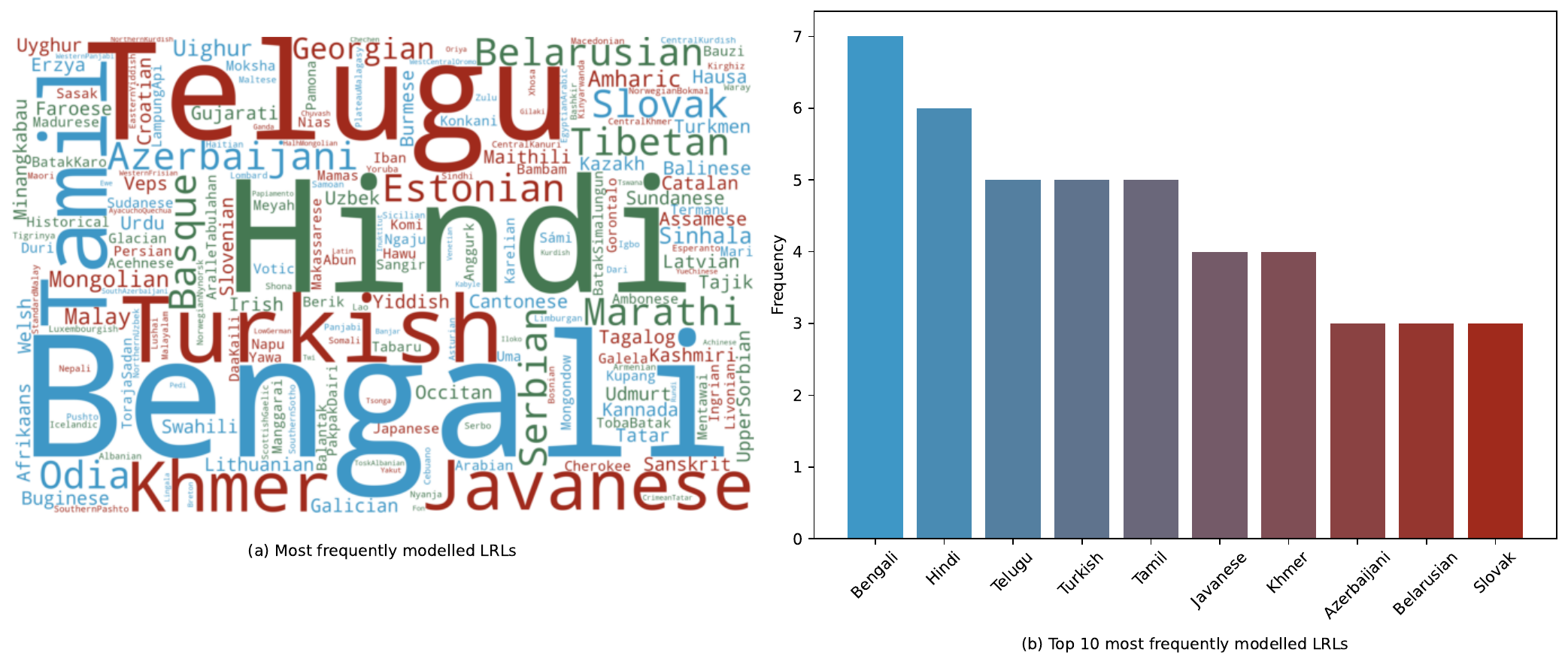}
\caption{Visualisation of the most frequently modelled low-resource languages (RQ1), shown as a word cloud and a bar chart of the top 10 LRLs.}
        \label{fig:lrlwordcloud}
    \end{figure}

\begin{figure}[htbp]
    \centering
    \includegraphics[width=0.8\textwidth]{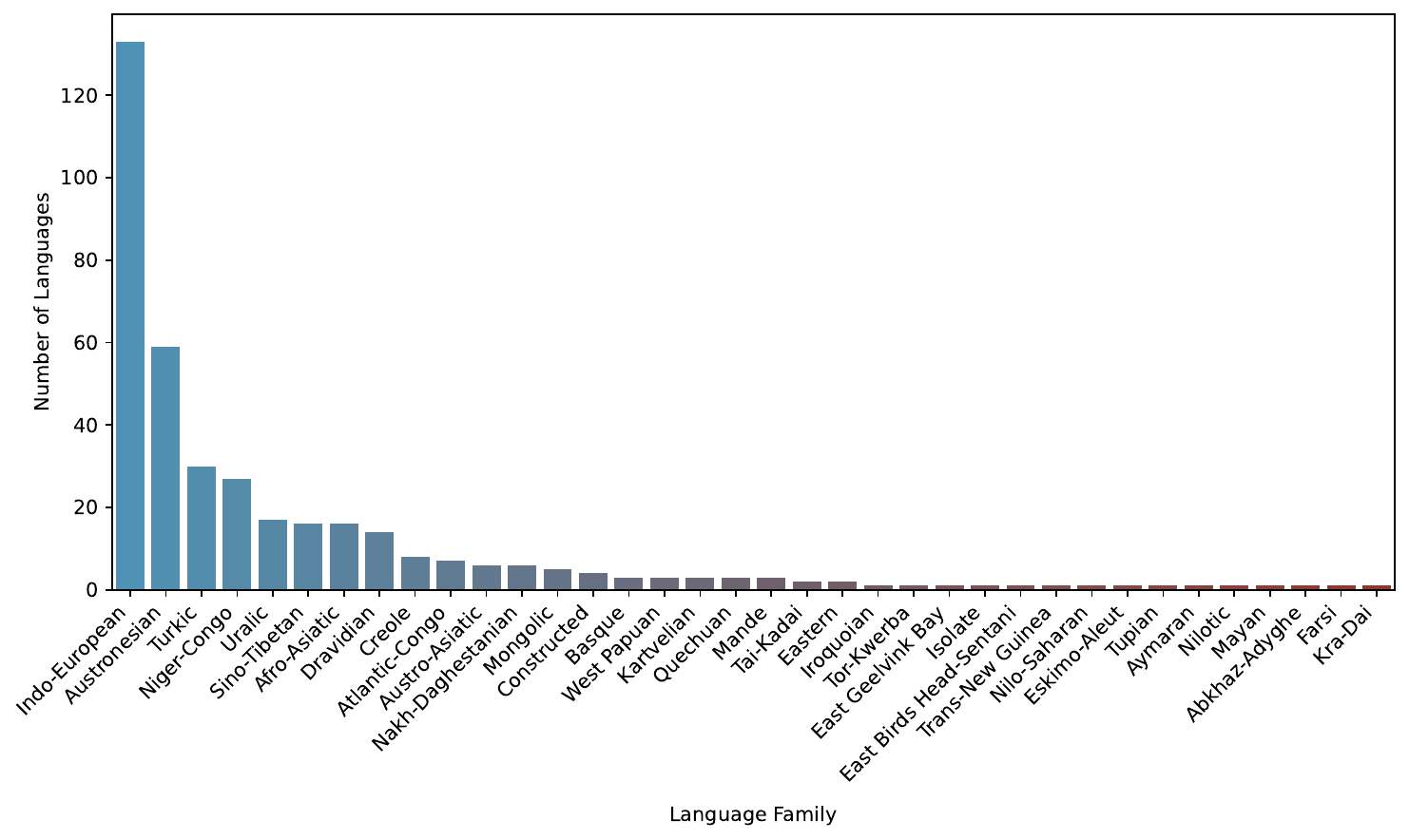}
    \caption{Distribution of the most frequently modelled low-resource language families by papers in this systematic review (RQ1).}
    \label{fig:lrlfamilies}
\end{figure}

    This review has identified a wide range of LRLs that have been modelled by other studies. A total of 295 unique LRLs have been trained spanning a variety of language generation tasks and different model architectures. Figure~\ref{fig:lrlwordcloud} showcases the most frequently modelled LRLs by studies examined in this review. Bengali is the most frequently modelled LRL, with 13\% of papers including it \cite{niyogi_paramanu_2024, noauthor_banglagpt_nodate,chang_when_2023,shafayat_benqa_2024,bhowmick_improving_2023,sorokin_ask_2022, agarwal_zero-shot_2022}. Hindi follows behind Bengali with the language appearing in 11\% of the studies \cite{niyogi_paramanu_2024,chang_when_2023,bhowmick_improving_2023, tanwar_translating_2020,gerz_language_2018,maimaiti_improving_2021}. Whereas, Tamil \cite{niyogi_paramanu_2024,chang_when_2023,gerz_language_2018,nissanka_exploring_2020}, Telugu \cite{niyogi_paramanu_2024,chang_when_2023,sorokin_ask_2022, tanwar_translating_2020,dandapat_iterative_2018} and Turkish \cite{chang_when_2023, agarwal_zero-shot_2022,gerz_language_2018, toraman_llamaturk_2024, acikgoz_bridging_2024,yirmibesoglu_morphologically_2023} are each modelled by 9\% of the papers, thus making them the third most commonly modelled LRLs. Although Turkish has been considered to be a high-resource language \cite{chang_when_2023}, some papers argue it is an LRL \cite{toraman_llamaturk_2024,yirmibesoglu_morphologically_2023}. Therefore we classify Turkish as an LRL. Moreover, Javanese \cite{chang_when_2023,gerz_language_2018, cahyawijaya_indonlg_2021, wongso_many--many_2023} and Khmer \cite{agarwal_zero-shot_2022, gerz_language_2018,mao_tuning_2024, pham_van_improving_2022} have both appeared in 7\% of the papers, with Azerbaijani \cite{chang_when_2023, pham_meta_2021, maimaiti_improving_2021}, Belarusian \cite{chang_when_2023, mao_tuning_2024, pham_meta_2021} and Slovak \cite{pham_meta_2021, chang_when_2023, gerz_language_2018} among the 16 languages that were modelled in 6\% of studies times. Figure~\ref{fig:lrlwordcloud} is a word cloud that displays more of the most frequently modelled LRLs. This word cloud highlights that LRLs such as Marathi \cite{niyogi_paramanu_2024, chang_when_2023, noauthor_bidirectional_nodate} and Odia \cite{niyogi_paramanu_2024,chang_when_2023, noauthor_generative_nodate} were also modelled somewhat frequently. Figure~\ref{fig:lrlfamilies} reveals aggregations whereby each language is assigned to a category according to its respective language family. We have identified languages from 39 different language families that have been modelled by studies in this review. LRLs belonging to the Indo-European language family consist of 35\% of the most frequently modelled LRLs. Austronesian languages are the second most commonly modelled LRL with a frequency of 15\%. Turkic and Niger-Congo-based languages are similarly represented with 8\% and 7\% frequency respectively. Uralic, Afro-Asiatic and Sino-Tibetan each represent 4\% of the LRL families modelled. Six remaining language families are represented with a frequency greater than 1\%. These language families include Creole, Atlantic-Congo, Austro-Asiatic and Nakh-Daghestanian at 2\%, with Mongolic and Constructed at 1\%. 

    \subsection{RQ2: What Technical Methods Have Been Used to Overcome Data Scarcity in Building Generative Language Models?}
\begin{figure}[htbp]
        \centering
        \includegraphics[width=0.8 \textwidth]{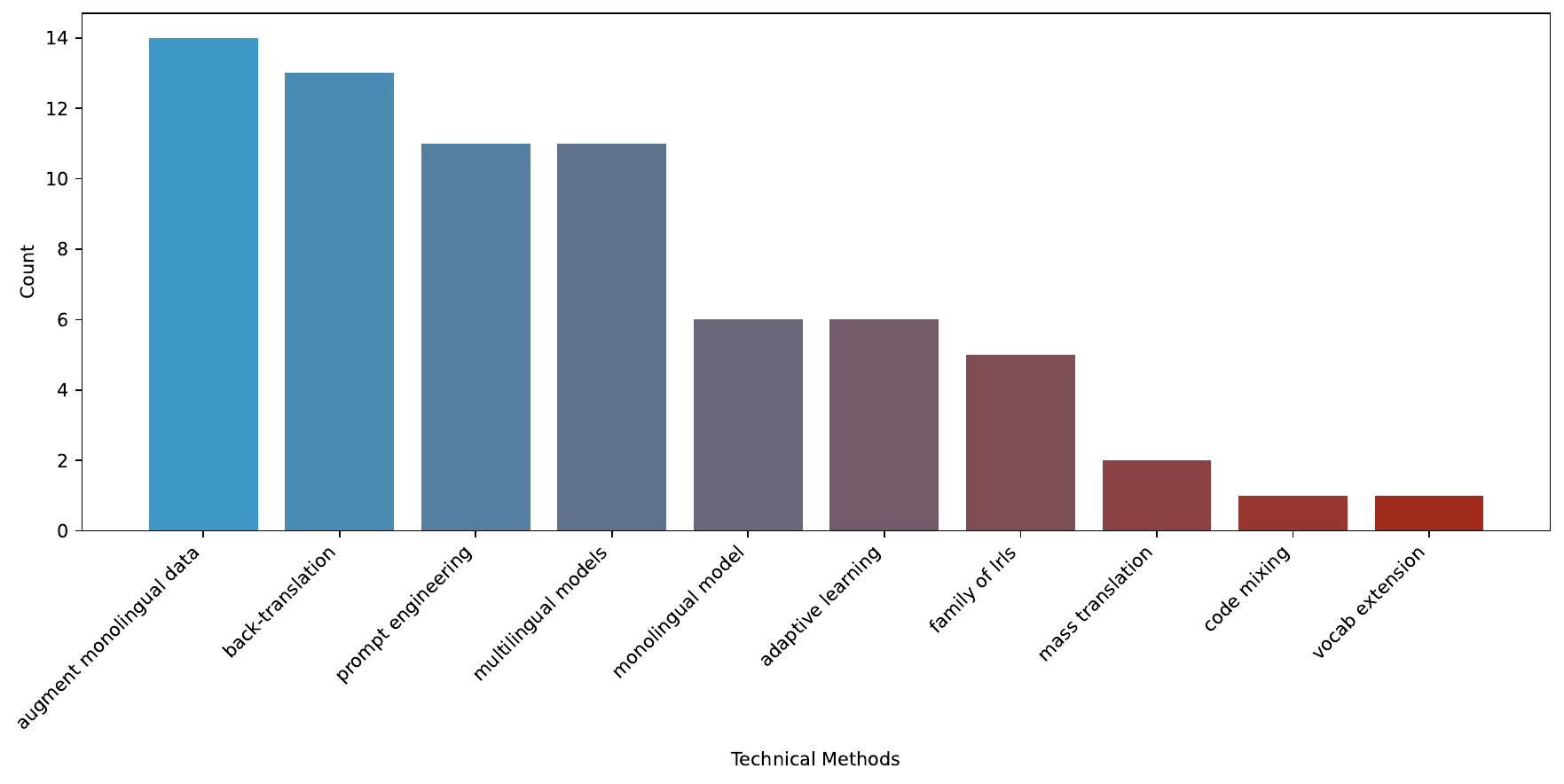}
        \caption{Distribution of the technical methods used to overcome data scarcity in building generative language models by frequency (RQ2). Note that these categories are not mutually exclusive, as multiple methods may be employed in a single study. }
        \label{fig:technicalmethods}
    \end{figure}

    A wide variety of technical methods have been used to overcome the issue of data scarcity associated with building generative language models for LRLs. We created a variety of categories that reflect the range of technical approaches undertaken by these papers that relate to filling the data gap required to create generative models. These categories are not mutually exclusive, as many papers employed multiple strategies in parallel. Each paper was assigned to one or more categories based on the techniques it discussed or implemented to tackle the issue of data scarcity. Figure~\ref{fig:technicalmethods} indicates that the most frequently mentioned technical approach was to perform some form of augmentation on monolingual data. The majority of the papers reference augmenting monolingual data, back-translation, building multilingual models, building monolingual models or prompting existing generative language models to approach the challenge of limited data for training. Table~\ref{tab:augmentation_techniques} provides an overview of how individual papers align with specific technical methods for overcoming data scarcity. The following sections discuss these methods in order of frequency, starting with the most commonly used approaches.  

        \subsubsection{Monolingual Text Data Augmentation}
            The application of various strategies for bolstering  monolingual text data occurs in 26\% of the studies. Such methods include generating new sentences for neural machine translation (NMT) by substituting low-frequency words with terms selected by an LSTM language model \cite{haque_recent_2021,shi_low-resource_2022}; constructing new sentences through the grammatical reordering of monolingual data using a reverse transcription grammar model \cite{wu_study_2023}; using a generative model combined with a grammar correction model to create grammatically correct synthetic data from the monolingual source data \cite{noauthor_qugan_nodate}; and duplicating original sentences in a monolingual corpus and using a model to paraphrase them \cite{shi_low-resource_2022, guo_automatically_2021, noauthor_paraphrase_nodate}. Other approaches include synthesising user-generated text for NMT by applying noise and stylistic changes to textual data \cite{marie_synthesizing_2020}; appending document-specific metadata and information to the document itself \cite{usui_translation_2023}; generating text data for NMT by turning the process of augmentation into an optimisation problem (SwitchOut) \cite{haque_recent_2021, shi_low-resource_2022, wang_switchout_2018}; and using POS tagging to make replacements based on the label of the term \cite{maimaiti_improving_2021,shi_low-resource_2022}. Additional strategies involve boosting training data by reordering monolingual target language text to align with the structure of another language \cite{haque_recent_2021}; incorporating knowledge at the word, phrase, and sentence levels to maximise the information gain from a monolingual corpus \cite{mi_multi-granularity_2024}; and using soft contextual data augmentation to randomly replace a word with a similar meaning \cite{haque_recent_2021, gao_soft_2019}.

\subsubsection{Back-Translation}
Back-translation is also frequently used as a technique for addressing data scarcity, appearing in 24\% of the studies
 \cite{nissanka_exploring_2020, dandapat_iterative_2018,yirmibesoglu_morphologically_2023,pham_van_improving_2022,pham_meta_2021,shi_low-resource_2022,marie_synthesizing_2020,wang_switchout_2018,hong_cantonmt_2024, zhang_chren_2020,berckmann_low-resource_2020, wang_survey_2022}. This approach involves boosting monolingual data by using a backward NMT model to translate a monolingual corpus from a target language to a source language. This process of translating to another language and back again can produce variations that bolster the source language dataset. Although this technique starts with monolingual data and shares some characteristics with monolingual augmentation methods, this review treats back-translation as a distinct category due to its reliance on machine translation systems. 

\subsubsection{Prompt Engineering}

A considerable portion of the studies, specifically 20\%, focused on addressing data scarcity by prompting existing generative language models. These existing models are examined based on their capacity to generate text for a wide range of LRLs. Several techniques have been proposed to improve generative language models. The programmer-interpreter approach iteratively refines a model’s output using a smaller language model for modifications \cite{li_improving_2024}. Another approach involves few-shot learning, where a model is taught unseen languages through examples and definitions \cite{noauthor_qasina_nodate}. Retrieval-augmented generation, which incorporates textbook-like material, enhances prompts and improves model output \cite{guo_teaching_2024}. Some studies evaluate existing generative models on their effectiveness in handling low-resource languages (LRLs) such as Vallader \cite{bendel_llegra_2024}, Odia \cite{noauthor_generative_nodate}, Faroese \cite{scalvini_evaluating_2024}, and Marathi \cite{noauthor_bidirectional_nodate}. Instruction tuning is another category within our review paper set that tackles the data disparity problem by employing a technical approach. This method involves providing clear instructions for language models that enable them to adapt to unseen tasks and LRLs. In addition to adding machine translation instructions, word alignments have also been used to enhance the fine-tuning process \cite{mao_tuning_2024}. The efficacy of instruction tuning has also been assessed for adapting generative large language models for LRLs \cite{toraman_llamaturk_2024}. We have determined that two papers use cross-lingual information sharing as a method to bolster the data available for LRL models. Both papers build a cross-lingual information system that is responsible for collecting information for languages with more available training data when the information does not exist in the desired target language \cite{sorokin_ask_2022,agarwal_zero-shot_2022}.

\subsubsection{Multilingual Models}
The category for multilingual models encapsulates all papers  that reference creating generative language models for more than a single language. 20\% of the studies were categorised for their use of multilingual models to address the issue of data scarcity when building generative models for LRLs. Several NMT models naturally build generative language models that generate translations for pairs of languages \cite{maimaiti_improving_2021,nissanka_exploring_2020,dandapat_iterative_2018, yirmibesoglu_morphologically_2023, pham_van_improving_2022, noauthor_bidirectional_nodate,pham_meta_2021,usui_translation_2023,wang_switchout_2018,mi_multi-granularity_2024,gao_soft_2019,hong_cantonmt_2024, zhang_chren_2020, berckmann_low-resource_2020,li_improving_2024, guo_teaching_2024, noauthor_english-arabic_nodate, ahmadnia_augmenting_2019, liu_low-resource_2022, noauthor_efficient_nodate, phan-vu_towards_2017}. In addition, several papers focused on building generative language models for a broader range of related languages. Some examples include a generative language model for 10 Indic languages \cite{niyogi_paramanu_2024}; a model trained on a variety of Uralic languages \cite{downey_targeted_2024}; a multilingual model trained on a mixture of 250 high-resource and low-resource languages \cite{chang_when_2023}; and IndoNLG, which serves as a benchmark for generative tasks involving Indonesian languages \cite{cahyawijaya_indonlg_2021}. Other notable multilingual models include a model paired with textbook-based prompts to teach LRLs to an existing language model \cite{guo_teaching_2024}; a neural approach for next-word prediction across languages \cite{gerz_language_2018}; and a model that leverages cross-lingual information to extract content from high-resource languages like English and present it in another language \cite{sorokin_ask_2022,agarwal_zero-shot_2022}. Additional work includes a translation model trained on a family of related Indonesian languages \cite{wongso_many--many_2023}, as well as a pre-trained language model designed for Chinese minority languages, such as Mongolian \cite{li_multi-tasking_2022}.

\subsubsection{Monolingual Models}

Papers tagged with the monolingual model category were almost as common as multilingual models in our systematic review, with the approach being adopted by 11\% of the papers. These papers encapsulate all studies that trained a generative language model for a particular language. This category differs from monolingual text data as some approaches involve building models for a single LRL without using any form of data augmentation. Some of these approaches include training a Turkish LLM via continual training on various models that were not explicitly trained on Turkish data, such as Mistral-7B \cite{acikgoz_bridging_2024}; developing a Tibetan question-answering (QA) corpus generation model that generates text and applies BERT for grammar correction \cite{noauthor_qugan_nodate}; building a sequence-to-sequence model for Slovenian as an LRL \cite{noauthor_frontiers_nodate}; creating an Indonesian religious QA model \cite{noauthor_qasina_nodate}; and training a Bangla-based generative pre-trained transformer \cite{noauthor_banglagpt_nodate}.

\subsubsection{Adaptive Learning}

Adaptive learning is featured as a common method for overcoming data scarcity, appearing in 11\% of the papers examined for this systematic review. This method relates to taking pre-existing language models and resuming the training phase of the model to expose it to more data. LLamaTurk is based on the open-source LLaMa-7b base and further trained on Turkish text data \cite{toraman_llamaturk_2024}. A similar approach is done for Turkish using Mistral-7b and GPT-2 as a base model \cite{acikgoz_bridging_2024}. Pre-trained models have also been fine-tuned to exploit cross-lingual capabilities for the Basque language \cite{otegi_conversational_2020}.

\subsubsection{Family of Low-Resource Languages}

There are a handful (9\%) of papers that propose the idea of training generative language models on families of LRLs as opposed to broad multilingual models with a diverse range of languages. Therefore this study includes this as a separate category. Some papers that build a language model trained on multiple related languages point out that a model built for a family of languages can perform better than monolingual models for those languages \cite{downey_targeted_2024,niyogi_paramanu_2024,cahyawijaya_indonlg_2021,wongso_many--many_2023}. 

\subsubsection{Mass Translation}

A pair (4\%) of papers were found to perform mass translation as a means to bridge the data gap when modelling LRLs. Tools like Google Translate API and GPT 3.5/4 have been used to generate parallel datasets for training QA models for Bengali \cite{shafayat_benqa_2024}. One survey paper references mass translation as a method for translating missing values within a parallel corpus \cite{haque_recent_2021}. 

\subsubsection{Miscellaneous}
 
A variety of studies have unique approaches that have been categorised on their own for tackling the issue of data scarcity when building generative language models. One of these approaches pertains to extending the vocabulary of a large language model by merging the vocabulary of an existing model with the vocabulary of an LRL such as Turkish \cite{toraman_llamaturk_2024}. Another study adopts the concept of code-mixing, whereby a sequence-to-sequence model is trained on parallel data to create sentences comprising a mix of languages \cite{bhowmick_improving_2023}. 

    \begin{table}[htbp]
    \centering
    \renewcommand{\arraystretch}{1.3}
    \setlength{\tabcolsep}{10pt}
\caption{Overview of techniques used to address data scarcity in low-resource language settings (RQ2), mapped to the corresponding papers that employed or proposed them.}
    \label{tab:augmentation_techniques}
    \begin{tabular}{|p{4cm}|p{10cm}|}
        \hline
        \textbf{Technique} & \textbf{Study} \\
                \hline
        Augment Monolingual Data & 
\cite{noauthor_qugan_nodate,usui_translation_2023,gao_soft_2019,marie_synthesizing_2020,maimaiti_improving_2021,guo_automatically_2021,wang_switchout_2018,noauthor_paraphrase_nodate,shafayat_benqa_2024,wu_study_2023,wang2021survey,haque_recent_2021,shi_low-resource_2022,zhang_neural_2024} \\
        \hline
        Back-Translation & \cite{zhang_chren_2020,marie_synthesizing_2020,ahmadnia_augmenting_2019,yirmibesoglu_morphologically_2023,dandapat_iterative_2018,wang_switchout_2018,pham_meta_2021,hong_cantonmt_2024,berckmann_low-resource_2020,pham_van_improving_2022,nissanka_exploring_2020,wang_survey_2022,shi_low-resource_2022} \\
        \hline
        Prompt Engineering & \cite{li_improving_2024,sorokin_ask_2022,agarwal_zero-shot_2022,scalvini_evaluating_2024,guo_teaching_2024,bendel_llegra_2024,mao_tuning_2024,toraman_llamaturk_2024,zhang_teaching_2024,noauthor_bidirectional_nodate,noauthor_generative_nodate} \\
        \hline
        Multilingual Models & \cite{bhowmick_improving_2023,cahyawijaya_indonlg_2021,wongso_many--many_2023,niyogi_paramanu_2024,downey_targeted_2024,chang_when_2023,tanwar_translating_2020,gerz_language_2018,liu_low-resource_2022,babaali_breaking_2024,li_multi-tasking_2022} \\
        \hline
        Monolingual Model & \cite{noauthor_qugan_nodate,noauthor_frontiers_nodate,mi_multi-granularity_2024,acikgoz_bridging_2024,noauthor_banglagpt_nodate,noauthor_qasina_nodate} \\
        \hline
        Adaptive Learning & \cite{otegi_conversational_2020,noauthor_efficient_nodate,noauthor_english-arabic_nodate,toraman_llamaturk_2024,acikgoz_bridging_2024,noauthor_qasina_nodate} \\
        \hline
        Family of LRLs & \cite{cahyawijaya_indonlg_2021,wongso_many--many_2023,niyogi_paramanu_2024,downey_targeted_2024,li_multi-tasking_2022} \\
        \hline
        Mass Translation & \cite{shafayat_benqa_2024,haque_recent_2021} \\
        \hline
        Code Mixing & \cite{bhowmick_improving_2023} \\
        \hline
        Vocabulary Extension & \cite{toraman_llamaturk_2024} \\
        \hline
    \end{tabular}
\end{table}

     \subsection{RQ3: What Monolingual Text Data Augmentation Techniques Have Been Used to Build Generative Language Models for Low-Resource Languages?}
\begin{figure}[htbp]
        \centering
        \includegraphics[width=0.8 \textwidth]{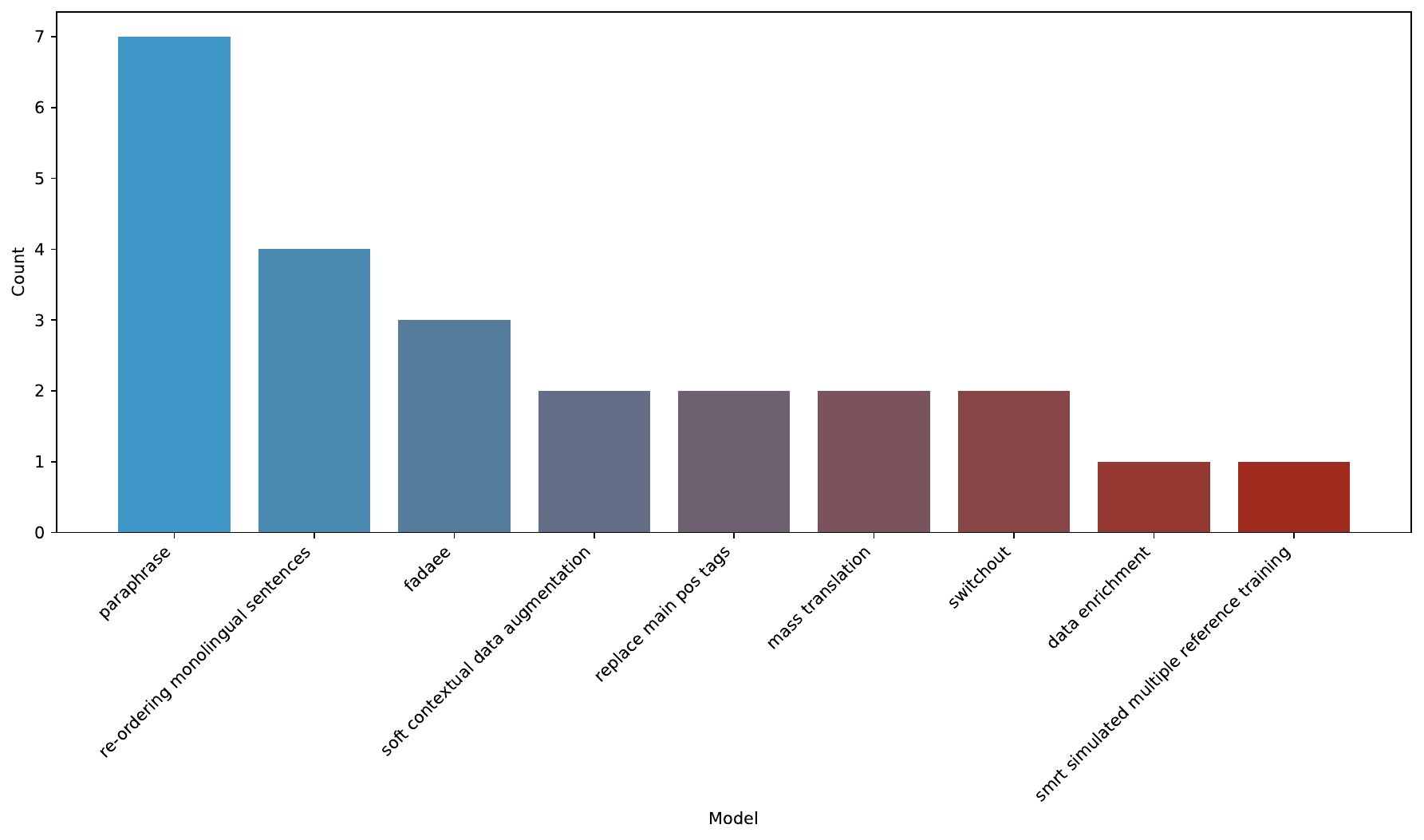}
        \caption{Distribution of the monolingual text data augmentation techniques identified in this systematic review by frequency (RQ3).}
        \label{fig:daugment}
    \end{figure}

    There are a variety of technical methods that have been adopted for addressing and minimising the issue of limited training data for LRLs that have been explored as part of RQ2. Given the utility of data augmentation in computer vision tasks \cite{shorten2019survey}, this systematic review dives deeper into applying similar synthesis processes for language modelling. Data augmentation techniques in this review relate to the papers that discuss synthesising new data from existing data for LRLs. We apply categories to each paper based on the type of text data augmentation used. The results of this section are displayed in Figure~\ref{fig:daugment}. The categories are organised into three broader groups based on the nature of the augmentation strategy: paraphrase-based techniques, grammatical transformation, and data enrichment. The following sections discuss these groups in order of frequency, starting with the most commonly used approaches. 
    
    \subsubsection{Paraphrasing}
    The creation of new monolingual sentences through paraphrasing is the most popular approach for augmenting textual training data for building generative language models for LRLs. This paraphrasing process itself differs from paper to paper. These differing approaches include building a model that can use a set of random terms and subsequently produce QA pairs \cite{noauthor_qugan_nodate}; enacting a variety of random word replacement strategies, such as rare-term replacement, simulated multiple reference training and soft context augmentation  \cite{haque_recent_2021, gao_soft_2019,shi_low-resource_2022,wang_switchout_2018, maimaiti_improving_2021}; and making part-of-speech (POS) informed word replacements, whereby specific word classes are replaced instead of random replacement \cite{maimaiti_improving_2021}. 

\subsubsection{Grammatical Transformation}
Grammar manipulation is used in a variety of papers in this systematic review to create synthetic sentences by applying various syntactic and structural changes to original training sentences. Some of these grammar manipulation processes include using a reverse transcription grammar model to generate alternative sentences for a given input sentence \cite{wu_study_2023} and introducing noise into existing training data to make it reflect user-generated text \cite{marie_synthesizing_2020}. 

\subsubsection{Data Enrichment}
One approach to data augmentation involved appending additional information to existing data. This information was in the form of metadata that encapsulated the title, access date, year and authors of a particular document \cite{usui_translation_2023}. 

    \subsection{RQ4: Which Sources Are Publishing in the Area of Low-Resource Language Modelling?}
\begin{figure}[htbp]
\centering
\includegraphics[width=0.8 \textwidth]{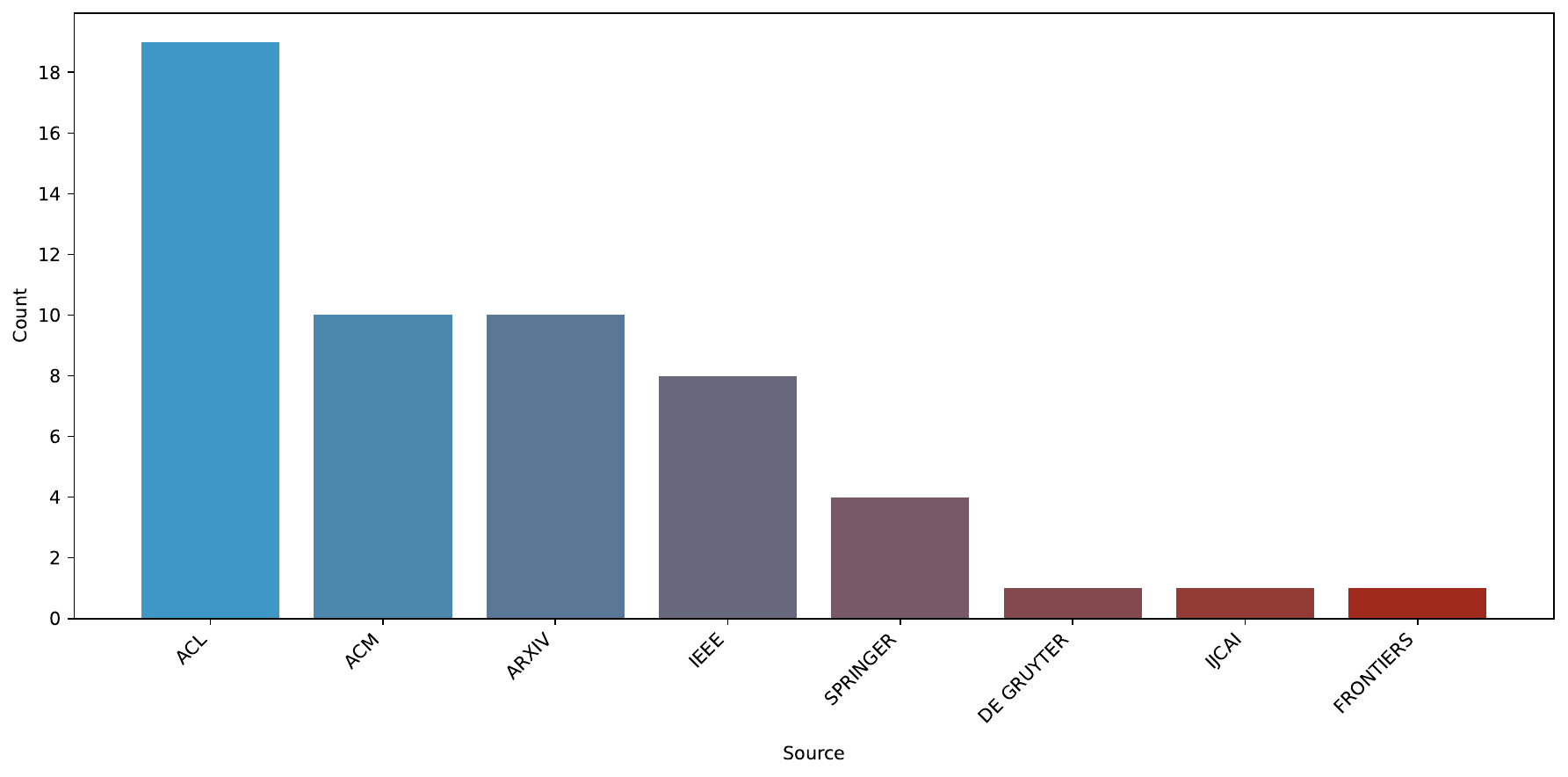}
\caption{Distribution of publishers by frequency (RQ4).}
\label{fig:publishers}
\end{figure}

    There are a variety of sources publishing in the area of LRL modelling. Figure~\ref{fig:publishers} indicates that the Association for Computational Linguistics (ACL) is the most active publisher in this area with 35.2\% of the papers belonging to their anthology \cite{sorokin_ask_2022,agarwal_zero-shot_2022,gerz_language_2018,dandapat_iterative_2018,cahyawijaya_indonlg_2021,mao_tuning_2024,noauthor_paraphrase_nodate,marie_synthesizing_2020,usui_translation_2023,wang_switchout_2018,gao_soft_2019,zhang_chren_2020, berckmann_low-resource_2020, li_improving_2024,liu_low-resource_2022,guo_teaching_2024, wang_survey_2022,otegi_conversational_2020, scalvini_evaluating_2024}. ACL represents 19 different publications spanning 13 conference proceedings, three workshops, and three journals. The Conference on Empirical Methods in Natural Language Processing \cite{zhang_chren_2020, cahyawijaya_indonlg_2021, wang_switchout_2018} and the Transaction of the Association for Computational Linguistics \cite{marie_synthesizing_2020, gerz_language_2018, wang_survey_2022} are the most frequently cited sources by ACL, both consisting of three papers. Furthermore, the International Conference on Language Resources and Evaluation \cite{scalvini_evaluating_2024, otegi_conversational_2020} and the Annual Meeting of the Association for Computational Linguistics \cite{gao_soft_2019, noauthor_paraphrase_nodate} are other conference papers by ACL that feature twice in this review. ACL also produces individual papers from nine other sources such as the Workshop on Multilingual Information Access \cite{agarwal_zero-shot_2022}; the Workshop on Technologies for Machine Translation of Low-resource Languages \cite{mao_tuning_2024}; the Conference on Machine Translation \cite{berckmann_low-resource_2020}; the North American Chapter of the Association for Computational Linguistics \cite{sorokin_ask_2022}; the Workshop on NLP for Similar Languages \cite{liu_low-resource_2022};  the International Conference on Computational Linguistics \cite{guo_teaching_2024}; the International Conference on Natural Language Processing for Digital Humanities \cite{usui_translation_2023}; the Annual Conference of the European Association for Machine Translation \cite{dandapat_iterative_2018}; the European Chapter of the ACL \cite{li_improving_2024}. 
    
\begin{figure}[htbp]
\centering
\includegraphics[width=0.8 \textwidth]{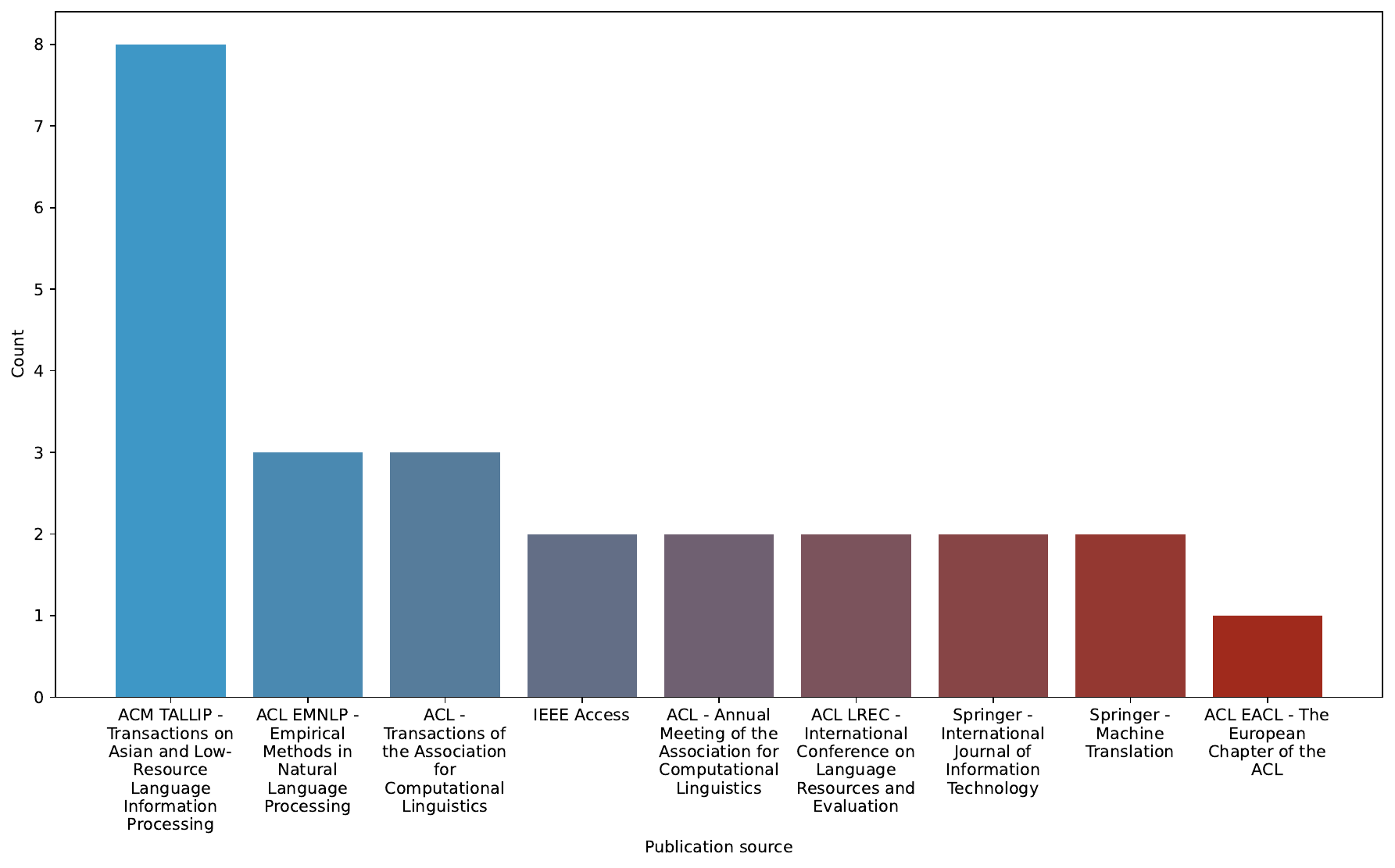}
\caption{Distribution of sources by frequency (RQ4).}
\label{fig:sources}
\end{figure}

    The Association for Computing Machinery (ACM) is the joint-second most frequently cited publisher in this field, accounting for 18.5\% of the papers \cite{bhowmick_improving_2023, yirmibesoglu_morphologically_2023, maimaiti_improving_2021, tanwar_translating_2020, pham_van_improving_2022, wu_study_2023, mi_multi-granularity_2024, shi_low-resource_2022, zhang_neural_2024, noauthor_efficient_nodate}. Although ACM is the joint-second most frequently cited publisher among the sources from this review, its journal publication Transactions on Asian and Low-resource Language Information Processing is the largest individual source of papers, as seen in Figure~\ref{fig:sources} \cite{bhowmick_improving_2023, yirmibesoglu_morphologically_2023, maimaiti_improving_2021, tanwar_translating_2020, mi_multi-granularity_2024, shi_low-resource_2022, zhang_neural_2024, noauthor_efficient_nodate}. ACM is also responsible for a single paper via the Symposium on Information and Communication Technology \cite{pham_van_improving_2022} and a single paper with the Asia Pacific Information Technology Conference \cite{wu_study_2023}. 
    
    Preprint papers from arXiv also constitute 18.5\% of the total dataset \cite{pham_meta_2021, niyogi_paramanu_2024, downey_targeted_2024, hong_cantonmt_2024, chang_when_2023, shafayat_benqa_2024, wang2021survey, toraman_llamaturk_2024, zhang_teaching_2024, acikgoz_bridging_2024}. There are no further aggregations possible for arXiv papers given that the platform allows users to upload and share literature without being published by traditional academic journals or undergoing formal peer review. Additionally, IEEE is the publisher of 14.8\% of the papers \cite{noauthor_qugan_nodate, wongso_many--many_2023, nissanka_exploring_2020, noauthor_banglagpt_nodate, noauthor_bidirectional_nodate, noauthor_english-arabic_nodate, noauthor_qasina_nodate, noauthor_generative_nodate}. IEEE Access is the only publication by IEEE that is cited more than once in this systematic review \cite{noauthor_qugan_nodate, wongso_many--many_2023}. However, IEEE is the source of six other publications: the International Conference on Information and Communication Technology for Sustainable Development \cite{noauthor_banglagpt_nodate}; the International Conference on Computer Systems and Applications \cite{noauthor_english-arabic_nodate}; the International Conference on Advances in ICT for Emerging Regions \cite{nissanka_exploring_2020}; the International Conference on Circuits \cite{noauthor_generative_nodate}; the International Conference of Advanced Informatics: Concept, Theory and Application \cite{noauthor_qasina_nodate}; the International Conference on Futuristic Technologies \cite{noauthor_bidirectional_nodate}. Moreover, Springer is the source of four papers consisting of 7.4\% of the papers \cite{bendel_llegra_2024, babaali_breaking_2024, li_multi-tasking_2022, haque_recent_2021}. Springer is the publisher of Machine Translation \cite{li_multi-tasking_2022, haque_recent_2021} and the International Journal of Information Technology \cite{bendel_llegra_2024, babaali_breaking_2024}, a pair of journals that are both cited twice in this review.  Finally, Frontiers \cite{noauthor_frontiers_nodate}, IJCAI \cite{guo_automatically_2021} and De Gruyter \cite{ahmadnia_augmenting_2019} are all credited as the publishers of a single paper respectively, bringing each of their contributions to 1.9\%. These publications include papers from: Frontiers in Artificial Intelligence \cite{noauthor_frontiers_nodate}; the International Joint Conference on Artificial Intelligence \cite{guo_automatically_2021}; and Open Computer Science \cite{ahmadnia_augmenting_2019}.
    
\begin{figure}[htbp]
    \centering
    \includegraphics[width=1 \textwidth]{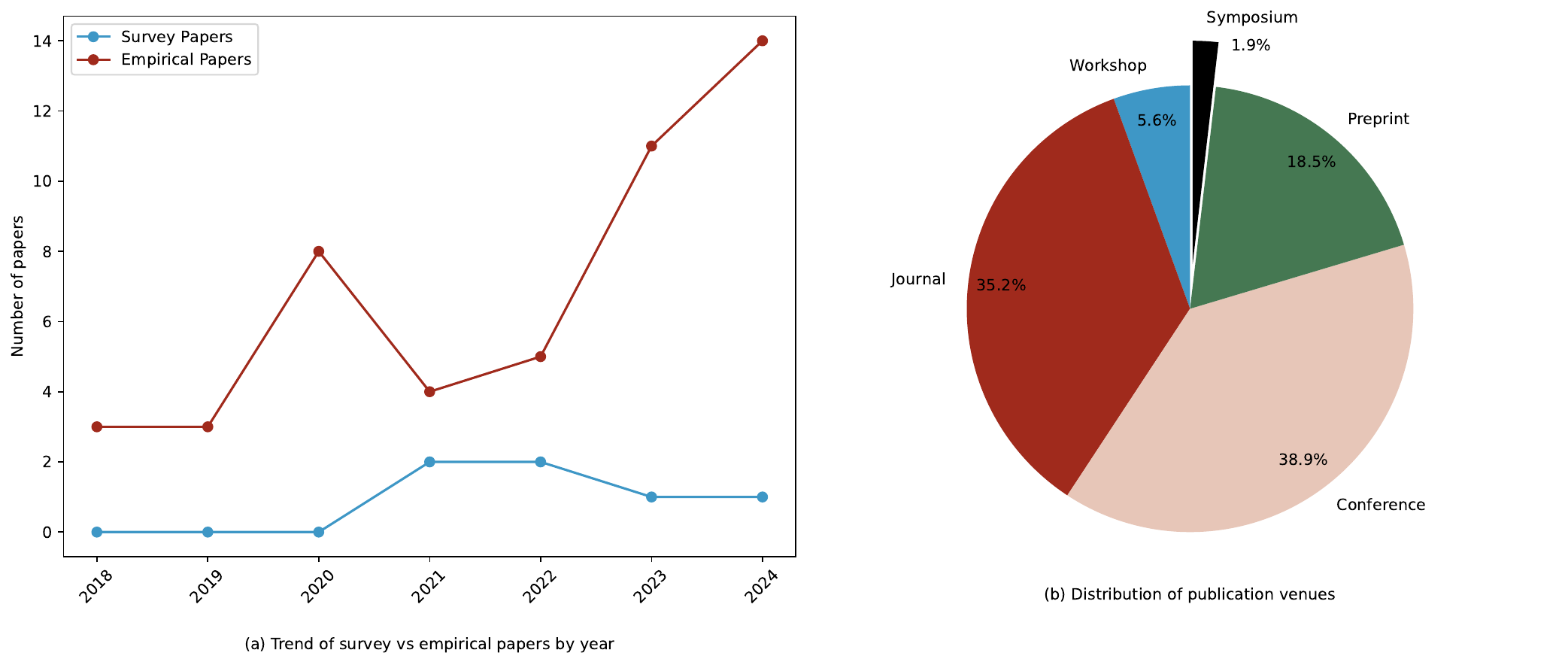}
\caption{Trend of survey vs empirical papers by year, and distribution of paper types by publication venue (e.g., conference, journal, symposium, etc.).}
    \label{fig:typeofpaper}
\end{figure}

\begin{figure}[htbp]
    \centering
    \includegraphics[width=0.8 \textwidth]{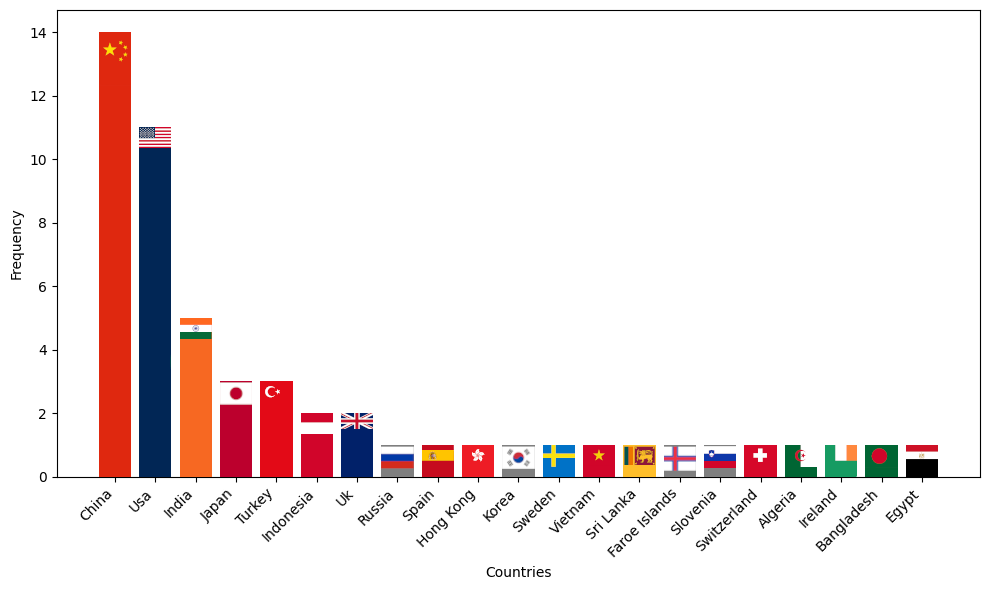}
    \caption{Distribution of publications by country.}
    \label{fig:countries}
\end{figure}

    The location of the research institution affiliated with the first author of each paper is represented in Figure~\ref{fig:countries}. China leads with the most publications, with the USA and India following behind. Figure~\ref{fig:core} indicates the CORE ranking distribution for conference papers and the Journal Impact Factor 2024 distribution for journal papers. 33.3\% of papers were not associated with a CORE-recognised conference. However, of the papers that are linked to a CORE-recognised event, 28.6\% of them were ranked A*. On the other hand, the majority (57.9\%) of journal publications were from Q3 journals. Figure~\ref{fig:typeofpaper} indicates that the number of empirical publications in this area has been increasing in recent years. 

    \subsection{RQ5: What Architectures Are Being Used to Produce Low-Resource Language Models?}
\begin{figure}[htbp]
        \centering
        \includegraphics[width=0.8 \textwidth]{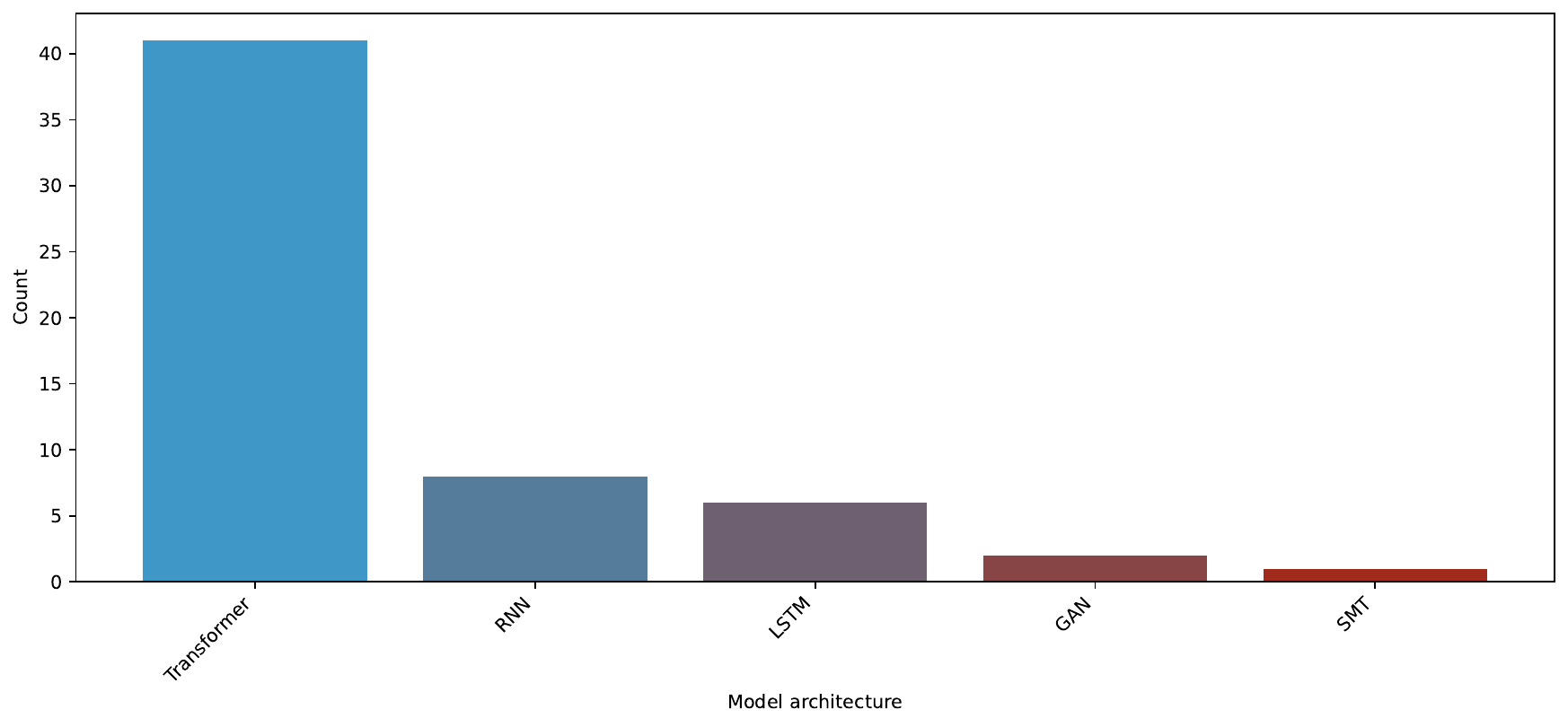}
        \caption{Distribution of the model architectures used by papers in this systematic review by frequency (RQ5). }
        \label{fig:architectures}
    \end{figure}

\begin{figure}[htbp]
        \centering
        \includegraphics[width=0.8 \textwidth]{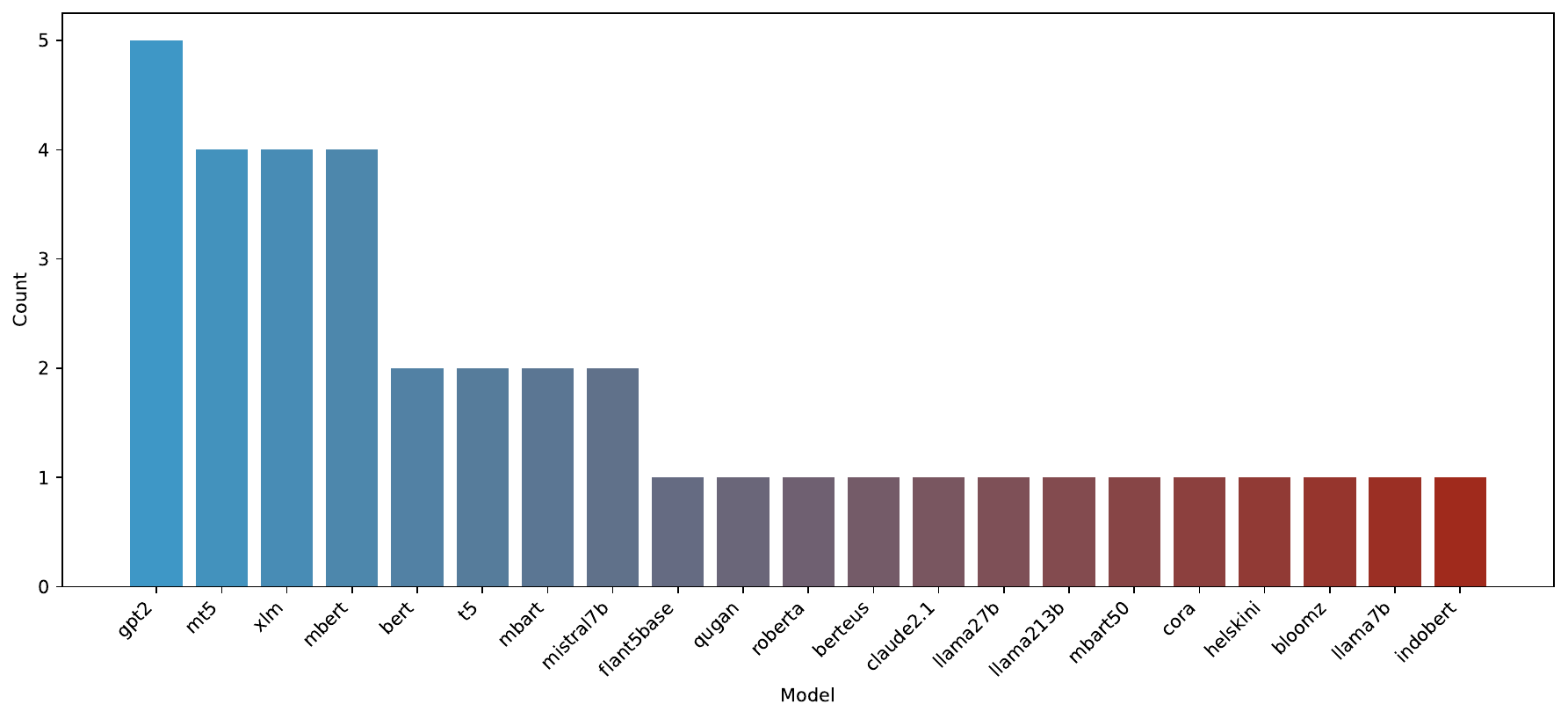}
        \caption{Model distribution by frequency within the systematic review (RQ5). Note: some studies report general architectures rather than specific models.}
        \label{fig:models}
    \end{figure}

    A diverse set of architectures have been adopted for LRL generation tasks. For this research question, we have assigned categories to each paper depending on both the general architecture of the models that are built and the specific models that are used. Figure~\ref{fig:architectures} indicates that the transformer architecture dominates as the architecture of choice for 76\% of the papers. In terms of the papers that model LRLs, the majority of them employ a transformer-based approach. Recurrent Neural Network-based implementations are the second most frequently adopted architecture for LRL modelling, with nine studies in this systematic review opting to utilise them. The Long Short-Term Memory (LSTM) categorisation applies to six papers. Given that LSTM-based approaches are inherently RNNs, it is worth highlighting that the categorisations are not mutually exclusive. 4\% of papers use Generative Adversarial Networks (GAN) for building LRL generation models. Additionally, a Gated Recurrent Unit (GRU) approach is implemented by one study, while another study employs a Statistical Machine Translation (SMT) architecture. In terms of the specific models being used to produce LRL models, Figure~\ref{fig:models} showcases the most frequently implemented models. GPT-2 appears most often in the studies with 11\% of papers adapting it for LRL modelling. XLM, MT5 and mBERT are each modelled by 7\% of studies. T5 is used by 6\% papers for LRL modelling. Furthermore, mBART and Mistral-7b are both used by 4\% of papers. The rest of the models are used once and include variations of LLama, BERT and BART models. 

    \subsection{RQ6: How Effective Have Methods for Overcoming Data Scarcity Been at Improving Performance for Different NLG Tasks?}
    Text data augmentations have been used by a variety of studies to improve the performance of language generation tasks for LRLs. Back-translation has been found to improve BLEU scores for translation tasks in comparison with no augmentation \cite{pham_meta_2021,hong_cantonmt_2024,berckmann_low-resource_2020,yirmibesoglu_morphologically_2023,nissanka_exploring_2020,dandapat_iterative_2018,pham_van_improving_2022}. This contrasts slightly with a minor BLEU improvement found by another study that uses back-translation to improve Cherokee-English machine translation \cite{zhang_chren_2020}. However, the study attributes the limited performance improvement to the lack of in-domain training data. Using a combination of synthetic code mixed and mixed-script with original data has been found to improve BLEU, METEOR and ROUGE scores for Indic language translation \cite{bhowmick_improving_2023}. Synthesising new text data via paraphrasing as a data augmentation technique has also been shown to increase the BLEU performance of translating from an LRL to English \cite{guo_automatically_2021}. Paraphrasing has improved BLEU results at both a character and word level for QA corpus generation for Tibetan \cite{noauthor_qugan_nodate}. Paraphrasing has also been used to improve BLEU scores by boosting data in low-resource tasks such as dialogue generation \cite{noauthor_paraphrase_nodate}. Reordering monolingual sentences as a form of text data augmentation has also yielded BLEU score improvements across a baseline of other data augmentation techniques \cite{wu_study_2023}. Similarly, synthesising user-generated text has been used as an augmentation technique to improve BLEU scores for NMT \cite{marie_synthesizing_2020}. Training a language model on a family of related languages is an effective data augmentation technique whereby a study has built an Indonesian language model that achieves similar results to state-of-the-art models with only ~20\% of the number of parameters \cite{cahyawijaya_indonlg_2021}. Another approach likens multilingual language modelling of related languages to a 33\% increase in the size of the monolingual dataset \cite{chang_when_2023}. Mass translation as a text data augmentation technique has also been found to improve COPA results for Bengali \cite{shafayat_benqa_2024}. Word-level augmentations have been successful at improving BLEU scores across other augmentation techniques for low-resource NMT \cite{maimaiti_improving_2021, wang_switchout_2018,gao_soft_2019}.
    
    \subsection{RQ7: How Scarce Is the Data in Studies Tackling Low-Resource Language Modelling?}
\begin{figure}[htbp]
        \centering
        \includegraphics[width=0.8 \textwidth]{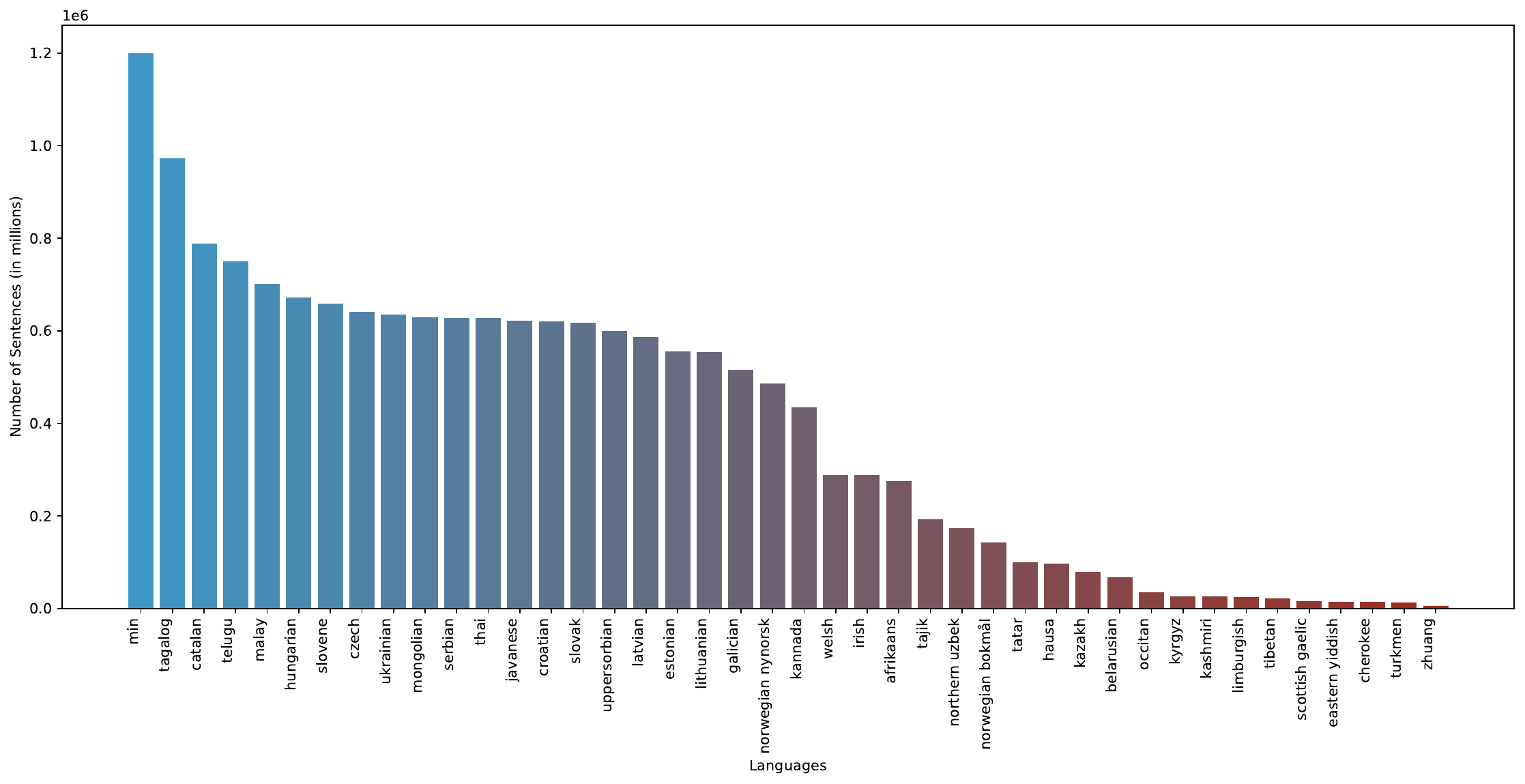}
        \caption{Distribution of sentence counts by language (RQ7).}
        \label{fig:sentences}
    \end{figure}

    As part of this review, we determine the quantity of data that has been used to model generative language models for LRLs. This process involved investigating and exploring the corpora that have been used to train models for LRLs. Given the ambiguity surrounding corpus statistics in many of the papers reviewed, this research question focuses on whether the study specifies the number of sentences used to train the LRL model. We found that three different datasets were used to train generative language models for Khmer \cite{agarwal_zero-shot_2022,mao_tuning_2024,pham_van_improving_2022} with data ranging from 70k-579k sentences. Figure~\ref{fig:sentences} highlights the languages with a single data source, where no additional sources are available. This visualisation indicates that Minnan has been modelled with the largest sentence count of 1.2m sentences \cite{gerz_language_2018}. From this aggregation, a count of 390k sentences is the average count for training language models for these LRLs. In several studies, data is quantified in terms of size in gigabytes rather than sentence count. For instance, Bengali models were trained on 26GB of data \cite{noauthor_banglagpt_nodate}, while Estonian, Turkish, and Indonesian models used 6.1GB \cite{downey_targeted_2024}, 180GB \cite{acikgoz_bridging_2024}, and 23.8GB \cite{cahyawijaya_indonlg_2021} of data, respectively.

    \subsection{RQ8: How Are the Models in These Studies Being Evaluated?}
\begin{figure}[htbp]
        \centering
        \includegraphics[width=0.8\textwidth]{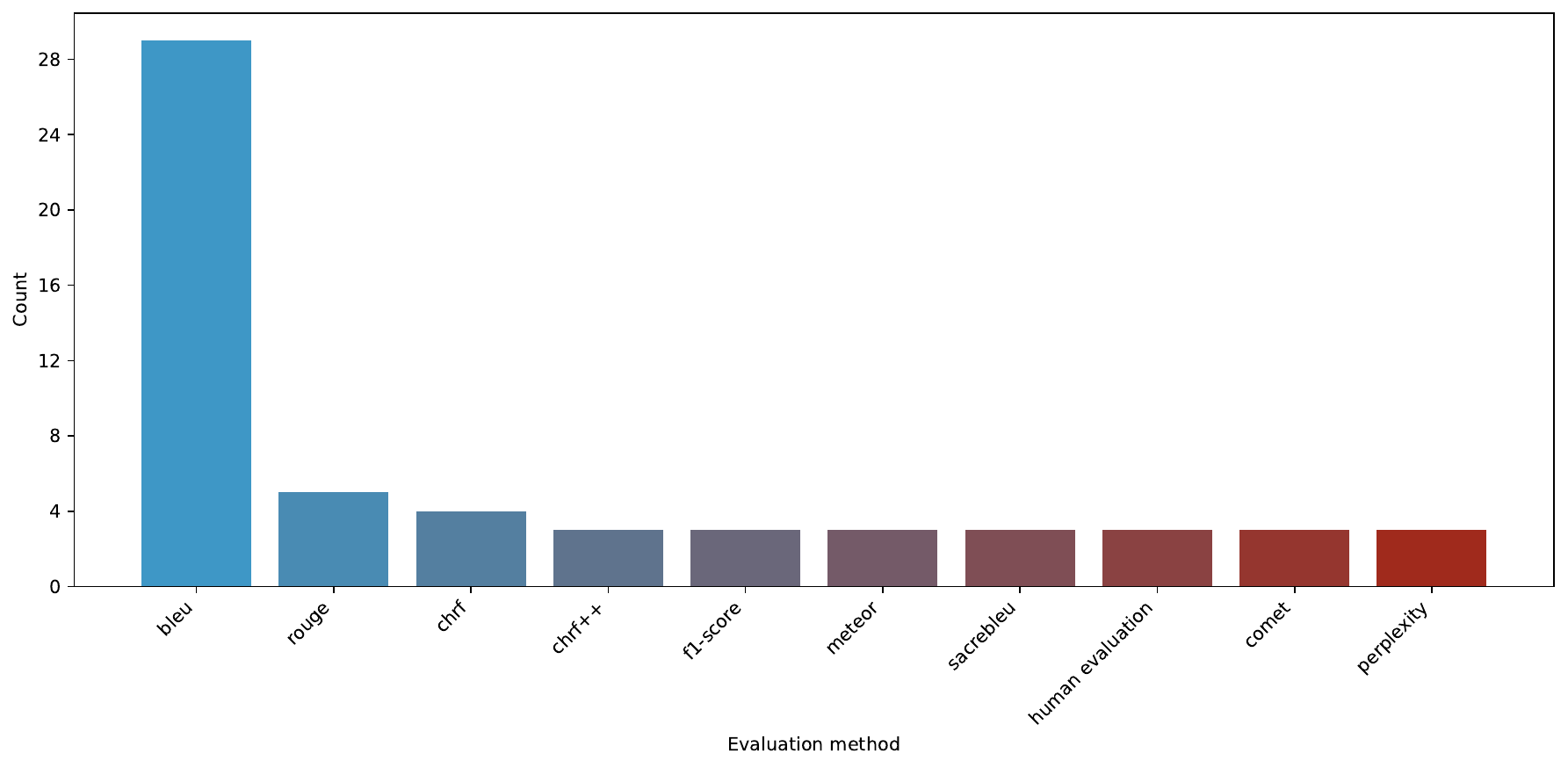}
        \caption{Distribution of evaluation method by frequency (RQ8).}
        \label{fig:eval}
    \end{figure}

Numerous evaluation methods have been adopted to assess the performance of NLG models for LRLs. Figure~\ref{fig:eval} indicates that BLEU dominates as the evaluation method of choice with 61\% of papers using it to gauge the efficacy of their models \cite{noauthor_banglagpt_nodate,bhowmick_improving_2023,tanwar_translating_2020,maimaiti_improving_2021,nissanka_exploring_2020,dandapat_iterative_2018,yirmibesoglu_morphologically_2023,cahyawijaya_indonlg_2021,mao_tuning_2024,pham_van_improving_2022,noauthor_bidirectional_nodate,pham_meta_2021,wu_study_2023,noauthor_qugan_nodate,guo_automatically_2021,noauthor_paraphrase_nodate,marie_synthesizing_2020,usui_translation_2023,wang_switchout_2018,mi_multi-granularity_2024,gao_soft_2019,hong_cantonmt_2024,zhang_chren_2020,berckmann_low-resource_2020,li_improving_2024,noauthor_english-arabic_nodate,ahmadnia_augmenting_2019,noauthor_efficient_nodate,phan-vu_towards_2017,li_multi-tasking_2022,otegi_conversational_2020,zhang_teaching_2024}. ROUGE is used by 9\% of studies \cite{bhowmick_improving_2023,noauthor_bidirectional_nodate,noauthor_english-arabic_nodate,li_multi-tasking_2022}, hence it is the second most frequently used evaluation method. Perplexity, chrF, and METEOR are each used in 7\% of the papers to evaluate their models \cite{noauthor_banglagpt_nodate, chang_when_2023,bhowmick_improving_2023,maimaiti_improving_2021,toraman_llamaturk_2024,noauthor_bidirectional_nodate,noauthor_english-arabic_nodate,li_multi-tasking_2022,scalvini_evaluating_2024,zhang_teaching_2024, vo_generative_2023}. Furthermore, five evaluation methods each appear in 6\% of the studies. These methods include human evaluation \cite{niyogi_paramanu_2024,dandapat_iterative_2018,hong_cantonmt_2024}; chrF++ \cite{mao_tuning_2024,li_improving_2024,guo_teaching_2024}, sacreBLEU \cite{wongso_many--many_2023, hong_cantonmt_2024,liu_low-resource_2022}; COMET \cite{mao_tuning_2024, hong_cantonmt_2024,guo_teaching_2024}; and F1-score \cite{sorokin_ask_2022,noauthor_qasina_nodate,otegi_conversational_2020}.

    \subsection{RQ9: What Is the NLG Task?}
\begin{figure}[htbp]
        \centering
        \includegraphics[width=0.8 \textwidth]{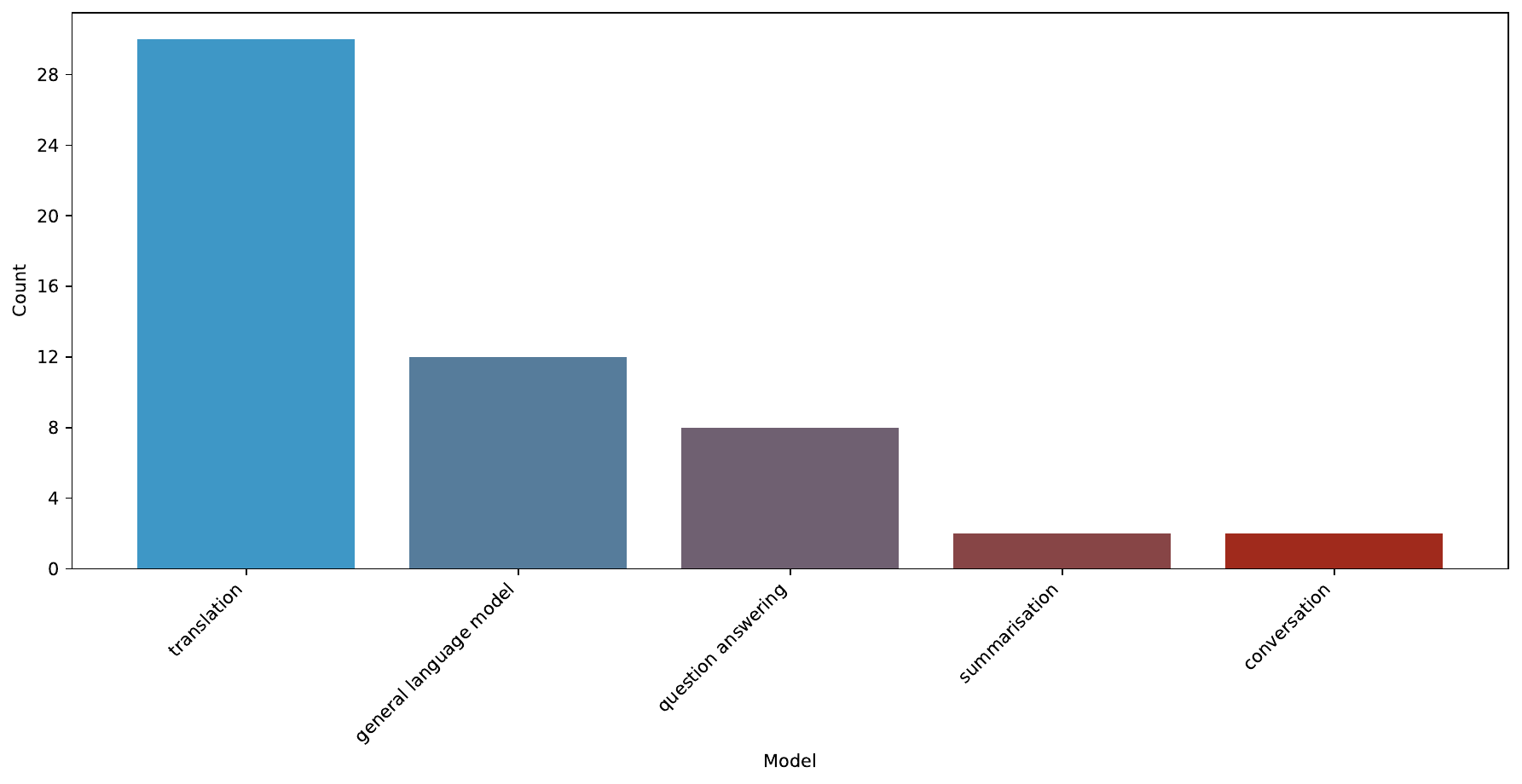}
            \caption{Distribution of language generation task by method (RQ9).}
        \label{fig:task}
    \end{figure}

A variety of generative models have been created for LRLs by studies. Figure~\ref{fig:task} showcases each NLG task by frequency. Given that some papers attempt to build models for multiple generative tasks, the categorisations are not mutually exclusive. Translation is the most popular generative task category, with 57\% of the studies focusing on the task \cite{bhowmick_improving_2023,agarwal_zero-shot_2022,gerz_language_2018,maimaiti_improving_2021,nissanka_exploring_2020,dandapat_iterative_2018,toraman_llamaturk_2024,acikgoz_bridging_2024,yirmibesoglu_morphologically_2023,cahyawijaya_indonlg_2021,wongso_many--many_2023,mao_tuning_2024,pham_van_improving_2022,noauthor_bidirectional_nodate,vo_generative_2023,downey_targeted_2024,pham_meta_2021,haque_recent_2021,shi_low-resource_2022, wu_study_2023,noauthor_qugan_nodate,guo_automatically_2021,wang_switchout_2018,li_improving_2024,noauthor_efficient_nodate,phan-vu_towards_2017,guo_teaching_2024,wang_survey_2022,otegi_conversational_2020,zhang_teaching_2024}. Furthermore, general language modelling is the second most popular language generation task  with 24\% of the papers tackling it \cite{niyogi_paramanu_2024,noauthor_banglagpt_nodate,chang_when_2023,gerz_language_2018,toraman_llamaturk_2024,acikgoz_bridging_2024,cahyawijaya_indonlg_2021,downey_targeted_2024,berckmann_low-resource_2020,noauthor_english-arabic_nodate,li_multi-tasking_2022,noauthor_frontiers_nodate}. We categorise general language modelling as research relating to building large pre-trained models as a base for various downstream tasks. Question-answering is also somewhat popular with 15\% of the papers building models for the task \cite{shafayat_benqa_2024,sorokin_ask_2022,agarwal_zero-shot_2022,noauthor_qugan_nodate,noauthor_qasina_nodate,otegi_conversational_2020,vo_generative_2023}. Summarisation \cite{noauthor_paraphrase_nodate} and dialogue generation models are also modelled by a single paper in this systematic review \cite{noauthor_english-arabic_nodate}.  

\section{Discussion}
    This section discusses the findings of the included studies, obtained through the data collection process, concerning the research sub-questions that inform the overall research objective of this systematic review. Each research sub-question is discussed in a separate subsection.

    \subsection{RQ1: What Low-Resource Languages Have Already Been Modelled by Other Studies?}
    The findings show a lack of a consistent and universal definition for what constitutes a low-resource language. In terms of low-resource being linked to data availability, some languages have relatively rich digital corpora compared with most underrepresented languages like Aymara. However, the concept of "low-resource" must also consider the availability of computational infrastructure for modelling underrepresented languages. The findings suggest that pre-training from scratch is rarely explored, with most papers fine-tuning models or prompting existing generative models. This lack of pre-training from scratch could indicate that there is also a technical limitation in researching low-resource languages. Additionally, research interest and the availability of researchers with technical expertise likely contributes to whether a language is low-resource. In other words, a language may remain “low-resource” not only due to data scarcity but also due to the absence of researchers or institutions capable of modelling it. That is to say that countries with stronger academic and technical ecosystems would have more opportunities to model their languages. Therefore, a language should be considered low-resource when it lacks any combination of data, computational power, or   researchers with relevant expertise.
    
    The predominance of Indo-European languages in the literature indicates a disparity among the LRLs themselves in terms of their representation. This is further evidenced by the 15  language families (such as the Aymaran and Tupian language families) that were only represented by a single language - versus the 133 Indo-European languages represented in the literature. This distribution could be attributed to region-specific disparities in research capacity. With most of the Indo-European languages belonging to European language branches, this suggests that more resources are available to researchers in Europe. The EU's commitment to the European Language Equality project and overall research and development (R\&D) funding (eg: via Horizon Europe) is indicative of strong support for Indo-European low-resource language technologies \cite{rehm2023european}. In contrast, the only Nilotic language represented \cite{chang_when_2023}, South Western Dinka, is spoken by an ethnic group in South Sudan \cite{lienhardt1958western}. To date, no official data has been published detailing South Sudan's R\&D spending as a percentage of GDP \cite{worldbank_rnd}. This highlights the geopolitical and economic factors shaping which languages are prioritised in language modelling.

    
    The lack of research on entire families of LRLs is particularly concerning given that the increasing dominance of AI tools could further increase language inequality in minority and underrepresented communities. With companies and governments alike looking to inject AI technology into their services \cite{dper_ai_guidelines2025, mckinsey_state_of_ai2025}, the lack of research for modelling entire families could contribute to technological marginalisation. Given that language models could be productivity-enhancing \cite{chatterjee2024impact}, their incompatibility with entire language families could introduce a new dimension of economic inequality. These LRL-agnostic technologies could widen the economic gap between regions that can fully leverage this technology and those that cannot. A lack of basic digital infrastructure like keyboards or unicode support for LRLs compounds the issue of language inequality \cite{zaugg2022digitally}. This lack of digital infrastructure presents a major challenge in sourcing sufficient data and building LRL models. 
    
    Given our findings surrounding the low-resource languages modelled by other papers, we recommend numerous areas for future research. Most notably, a more comprehensive dataset of LRLs should be created that accounts for the availability of data, computational resources and academic structures/technical expertise, in line with our proposed definition of "low-resource". This type of research could arm LRL NLP researchers with a strong foundation on the opportunities and challenges facing a given LRL. Furthermore, our review has found that there are benefits to building LRL models for a family of related languages. Tackling LRL modelling from a language family perspective could boost the representation of underrepresented languages. In addition, a language family-based approach would allow communities to pool computational resources and technical expertise. Therefore, further research is required on family-based language modelling as a sustainable and inclusive strategy for addressing language inequality in NLP.
    
    \subsection{RQ2: What Technical Methods Have Been Used to Overcome Data Scarcity Building Generative Language Models?}
    Similarly to the results section for RQ2, the following discussion and meta-analysis is divided by technical methods used to overcome data scarcity in order of frequency, starting with the most commonly used approaches. 
    
\subsubsection{Monolingual Text Data Augmentation}

The process of synthesising new data from monolingual text data is the most frequently used approach for overcoming data scarcity. While techniques ranging from simple word replacements to sophisticated paraphrasing and grammatical reordering are used, the effectiveness of these methods varies significantly depending on resource availability and task specificity.

The use of word-level replacements such as synonym and low-frequency term substitutions on existing data is limited by the replacement strategy. Even going beyond simple synonym replacement and using neural approaches like BERT to contextually replace terms can still corrupt the original meaning of the text data or break sentence-level agreement. Improvements such as using POS tags to substitute terms using a paraphrase table could work for some LRLs. Yet, it could be challenging to apply this to LRLs with a complex morphological structure such as Gaeilge, where altering a term in a sentence could corrupt the grammatical agreement of the sentence \cite{legate1999morphosyntax}. We propose that performing POS-informed replacements with a contextualised model like BERT could offer a reduction in sentence-level errors for LRLs where specific POS tags do not affect agreement. One of the main flaws of word-level replacements is that this method does not introduce or expose a model to new semantic content. While such methods may increase lexical diversity or syntactic variation, they do not expose the model to new facts, knowledge, or domain-relevant information. As a result, these techniques may be insufficient for tasks that require broader world knowledge or domain-specific understanding, such as QA. Therefore, word-level replacements could suit augmenting training data for translation tasks over other generative tasks like dialogue generation or abstractive summarisation. 

Alternatively, sentence-level transformations, often using some form of grammar correction, are also common methods for overcoming data scarcity. Similarly to the benefits of word-level augmentations, sentence-level transformations can increase sentence diversity in the training data. However, sentence-level transformations differ as they aim to preserve the meaning and grammatical features of text. While neural grammar correction strategies have been used to improve the quality of generated text \cite{noauthor_qugan_nodate}, the improvement hinges heavily on the availability and efficacy of the grammar correction tools. Nonetheless, effective grammar correction tools could enhance both word and sentence-level transformations. Reducing the likelihood of grammatical agreement errors in generated text via these kinds of grammar tools could make these synthetic sentences more suitable as training data for generative tasks like QA. Alternatively, other papers have experimented with reordering the sentences to match the syntax of target languages. However, using a reverse transcription model to generate new sentences through grammatical reordering of the original sentences could fail for languages with strict word order constraints. Given that some languages are more flexible in terms of their word order \cite{dryer2007word}, adopting a model to generate grammatically correct alternative sentences could enrich a dataset of LRL text by increasing the diversity of language. This could be useful for languages with dialects that alter the word order as it could produce dialectal variations from the original sentences. Therefore, these grammatical-based sentence-level transformations could be beneficial for translation tasks and could be explored for translations between dialects.

While these substitution methods can benefit LRL modelling, sentence-level transformations with grammar correction capabilities offer the most potential across a wider range of generative tasks. Future work should explore task-aligned augmentation pipelines that go beyond surface-level variation and actively incorporate new semantic information into training data.  

\subsubsection{Back-Translation}
Given that back-translation is the second most frequently used data augmentation technique and translation is the most common NLG task, this suggests that translation is of particular importance to minority language communities. The widespread use of back-translation underscores the extent of data scarcity and institutional neglect facing LRLs in global language technology development. 

Similarly to other translation-based data augmentation techniques, the efficacy of back-translation is heavily dependent on the quality and performance of the translation tools that are being used. While back-translation has improved translation performance for modelling a variety of LRLs \cite{berckmann_low-resource_2020,pham_van_improving_2022,nissanka_exploring_2020}, this process could consequently produce synthetic text data that disrupts the original semantics and grammar \cite{park-etal-2024-translation}. We propose that sentence-level grammar correction methods could alleviate grammatical errors produced via back-translation. Given that modelling groups of related languages can improve LRL modelling performance \cite{cahyawijaya_indonlg_2021}, back-translation could also be used to augment data across a variety of LRLs. In particular,  this could benefit modelling dialects of LRLs that differ syntactically from the standard form of a language. We propose that translating dialectal Irish to English, and then back to standard Irish, could yield parallel sentence pairs suitable for training dialect-aware generative models.

Overall, back-translation can benefit translation tasks but semantics can be lost in translation, suggesting it would not generalise well for augmenting summarisation datasets where translation errors could contribute to inaccuracies in output. Future research should explore the role of back-translation in augmenting datasets for dialect modelling and language family-level training and investigate hybrid pipelines that combine translation with grammar correction to ensure data quality across tasks.

\subsubsection{Multilingual Models}

Building multilingual models is an approach adopted by many of the papers in this study to address the lack of training data for LRLs. However, its more frequent use over monolingual modelling may reflect the need for data pooling across different languages to accurately model LRLs.

For LRLs, the cross-lingual transfer gained by modelling different languages together can improve performance across different tasks \cite{niyogi_paramanu_2024,chang_when_2023,downey_targeted_2024,cahyawijaya_indonlg_2021,guo_teaching_2024,sorokin_ask_2022,gerz_language_2018,agarwal_zero-shot_2022,wongso_many--many_2023,li_multi-tasking_2022}. Given that building multilingual models has been likened to augmenting the training dataset by 33\% \cite{cahyawijaya_indonlg_2021}, it could be described as an implicit form of data augmentation. Although synthetic data is not created, this inclusion of multilingual data exposes models to a broader set of language instances. As a result, this suggests more work should be done to pool text data resources between related languages for optimal model training. 

However, some evidence shows that some smaller multilingual models have been outperformed by their monolingual counterparts \cite{noauthor_frontiers_nodate}, suggesting that multilingual models perform better as they scale. This tradeoff could highlight the importance of selecting compatible language groups and having sufficient data between the different languages to avoid overfitting. Optimising language groupings remains an open research challenge.

Similarly, the use of code mixed sentences to generate translations could be extended to model languages or dialects of languages that often reflect authentic language practices in bilingual communities. While this could create seemingly nonsensical data, code-mixing could accurately reflect some communities such as Irish people, as they often inject English into their Irish conversations and vice versa \cite{laoire_irish-english_2016}. Therefore, code-mixing could be a valid method of augmenting data for a pair of languages that are frequently used in conjunction with each other.   

In addition to structural transfer, several studies have used cross-lingual knowledge transfer to exploit information across different language sources \cite{sorokin_ask_2022,agarwal_zero-shot_2022}. Collating information from higher-resourced languages to generate more relevant text in a target language presents some opportunities and challenges. Previous critiques of monolingual and back-translation augmentations target their inability to synthesise new semantic content. Therefore, this cross-lingual knowledge-sharing approach could expose LRL models to more diverse information that would otherwise not exist in the target language dataset. For example, some LRLs may lack news services that provide coverage of global affairs in their language, preventing this information from existing in a dataset for target LRLs. Consequently, facilitating cross-lingual information sharing could help bridge this gap, as source information could be rephrased in a target LRL. This would ensure that speakers of these languages have access to a broader range of knowledge not written explicitly in the target language. 

Future work could involve investigating the use of cross-lingual transfer to build generative models for LRL communities that code-mix between languages. This could be particularly beneficial for LRL languages that are code mixed with more resourced languages like Irish and English, as this could alleviate the data scarcity issue. In addition, more research is required on grouping data resources across related languages so that LRL modelling can benefit from cross-lingual knowledge transfer.  

\subsubsection{Prompt Engineering}
Prompt engineering has appeared as a method for overcoming data scarcity in several papers. The approach is likely popular given that it is less resource intensive in comparison with model fine-tuning or pre-training from scratch. Its popularity may reflect growing interest in evaluating whether existing generative tools can effectively support LRLs. While prompting existing models might be more accessible, it has its downsides. No matter the prompting approach, generation performance hinges on the multilingualism of the language model being prompted. If a prompted model has been exposed to relatively little data from a language or if the model lacks adequate tokenisation support for a language’s script, then it is likely that the model would struggle to generate accurate text. 

Two approaches inject additional information into prompts about the target language and provide examples of translations to guide the generation \cite{guo_teaching_2024,zhang_teaching_2024}. Although enriching prompts with textbook-like information can improve contextual grounding, the effectiveness of such methods is limited by the model’s tendency to ignore or truncate long or complex prompts. Often, lengthy prompts with too much information can result in prompt sections being forgotten or omitted \cite{yang2025prompts}. This would be a challenge for the approach of using textbook-like information. Ensuring that a prompted model follows the injected instructions is an open challenge across the field of prompt engineering. 

We recommend that future work should evaluate existing language models for their capacity to generate text in LRLs. Prompting existing models could serve as a baseline comparison for future fine-tuned or pre-trained models. Additionally, more work is required across the field of language modelling to develop models and decoding strategies that more reliably attend to and follow prompt instructions, especially in multilingual and low-resource settings.

\subsubsection{Monolingual Models}
While monolingual modelling is used by some studies, it is not as popular as multilingual modelling. The difference in popularity could suggest that there is growing interest in multilingual modelling as a method to tackle generative LRL modelling. 

This process of training models on monolingual data is often done by exploiting adaptive learning techniques. While one study that fine-tuned Mistral on Turkish data improves accuracy across reasoning and QA tasks \cite{acikgoz_bridging_2024}, there are several assumptions required for this method to work. Tokenisation support is essential; without it, a model may fail to represent a language’s syntax or morphology accurately. Additionally, a large language model with billions of parameters would require a significant amount of text data to avoid overfitting during the fine-tuning process. Previously discussed augmentation techniques such as monolingual text data augmentations could help with this challenge by producing synthetic data.  Therefore, fine-tuning an existing model could remove the need to pre-train a model from scratch, but the issue of data scarcity ultimately remains an open challenge. 

Some evidence shows that small monolingual T5 models can outperform small multilingual models for LRLs \cite{noauthor_frontiers_nodate}. While smaller monolingual models can outperform smaller multilingual ones, multilingual models tend to outperform as model size and available training data increase. This suggests that monolingual models could be suitable for modelling extremely low-resource languages, whereas larger multilingual models could perform better where there is considerable data available for a group of related languages. The stronger performance of multilingual models at scale may explain their growing popularity for LRL modelling. Given that some studies do not report appropriate baseline comparisons \cite{noauthor_banglagpt_nodate}, monolingual models could be a suitable baseline benchmark for LRL feasibility. If a model trained on monolingual data can perform well, the resource gap for the language may not be as large as perceived. 

We recommend that future work should focus on using monolingual modelling as a baseline benchmark for whether a language is as low-resourced as perceived. The prevalence of adaptive learning and the data requirements necessitate the collection of more data for LRLs to avoid overfitting. Future research could also investigate the tipping point at which multilingual models begin to consistently outperform monolingual models for LRLs, depending on model size, data availability, and language similarity.

\subsubsection{Adaptive Learning}

Adaptive learning has been leveraged by several studies to model LRLs. While two papers extended the training of pre-trained models by exposing them to additional Turkish data \cite{toraman_llamaturk_2024,acikgoz_bridging_2024}, this approach is prone to significant challenges. As explored in the monolingual modelling section, further training an existing pre-trained model allows it to benefit from transfer knowledge but it still needs to train on a significant amount of text data for models with billions of parameters. In addition to needing a large amount of data to avoid overfitting, the pre-trained models must support the tokens required to represent a given LRL. However, some evidence shows that resuming training with new raw text and extending the vocabulary size does not have an impact on performance \cite{toraman_llamaturk_2024}. Further research could be done to confirm these findings for other LRLs. 

One study observed that resuming training can trigger catastrophic forgetting of the primary language knowledge of multilingual models \cite{acikgoz_bridging_2024}. This could suggest that the data used to continue training needs to be representative of the existing split of languages learned by the model. On the other hand, if monolingual modelling is the objective then catastrophic forgetting could be ideal as a multilingual model adapts to a target language during training. The same study indicates that large models like Gemma7B still perform better without fine-tuning. This could suggest that collating a larger more representative dataset and building models with more parameters could be more beneficial for modelling multiple LRLs at once.

An open challenge would be to explore ways of extending models to account for new changes in language. Without ongoing updates, models will not capture emerging language, such as slang or cultural references (e.g., a model trained before 2024 would not understand the cultural context behind the phrase “brat summer”, popularised by British singer Charli XCX \cite{guardian_brat_summer_2024}). While adaptive learning can address data scarcity in the short term, future work should investigate how it can also support long-term model evolution, particularly for low-resource languages.

\subsubsection{Family of Low-Resource Languages}
Although families of LRLs are used less than both multilingual and monolingual modelling, the approach is heavily linked to multilingual modelling. The difference in popularity may reflect that not all multilingual modelling approaches are explored using related languages. However, the findings suggesting that multilingual modelling can yield performance gains equivalent to a 33\% increase in training data, should be further investigated in the context of related LRLs \cite{chang_when_2023}. This improvement could be enhanced by selecting similar or maximally related languages for multilingual modelling. This argument draws on the evidence that shows more resourced languages as well as LRLs can benefit from being modelled with related languages \cite{downey_targeted_2024, cahyawijaya_indonlg_2021}. The reduction in required parameters through modelling families of related languages \cite{cahyawijaya_indonlg_2021} should also be investigated further as reducing model sizes could lower both the cost and environmental impact of LRL modelling.

We agree that it could be infeasible to model some extremely LRLs (such as Komi or Moksha) without exploiting multilingualism and transfer learning across languages \cite{downey_targeted_2024}. Therefore, this necessitates further research focusing on pooling dataset resources across related linguistic communities. Future work should also investigate modelling families of LRLs as a way of reducing model sizes. This branch of research could promote environmentally friendly methods of solving the data scarcity issue for LRLs.

\subsubsection{Mass Translation}
Using tools like Google Translate API and GPT 3.5/4 as a form of mass translation is one of the least popular methods used for overcoming LRL data scarcity. Its lack of popularity is likely linked to the same pitfalls facing other previously discussed translation-based approaches, like back-translation. The major pitfall relates to the constraints of existing translation tools. Augmenting a monolingual dataset by leveraging a translation service to bulk translate from a source language to a target language limits the performance of subsequent models. In other words, any translation errors  would propagate through to any models trained on the synthetic data. While one study uses this approach to build QA systems for Bengali \cite{shafayat_benqa_2024}, these results would be constrained by the translation service's capacity to translate Bengali text data. For a task like QA where factual information could be queried, losing semantic content during translation could produce corrupted information. 

Although mass translation is not a widely favoured approach to addressing data scarcity, exploring the use of translated data to train generative models could still serve as a useful baseline for comparison with models trained on authentic text. Future work could explore translation methods that focus on preserving semantic content during translation. Ensuring translation systems do not distort factual information may result in a more useful approach for augmenting LRL data.

\subsubsection{Miscellaneous}

While code-mixing could be considered a form of multilingual model, it is also tagged with the miscellaneous category as it differs in its output. Code-mixing could model languages or dialects that use terms from multiple languages interchangeably. Given that Irish speakers often inject English into their Irish phrases and vice versa \cite{laoire_irish-english_2016}, future work could investigate using code-mixing to augment data for representing bilingual communities. 

Additionally, extending the vocabulary of existing pre-trained models is related to adaptive learning but it is also tagged with the miscellaneous category. Given that pre-training from scratch can be costly and time-consuming, adapting existing models that may not have tokenisation support for LRL scripts is a challenge. However, some evidence highlights that extending the vocabulary size of a model along with continuing training on new text might not have an impact on performance \cite{toraman_llamaturk_2024}. Although this study did not find any evidence of performance changes, further research could investigate the impact of the quantity of new data when extending the vocabulary size on model performance.     

\subsection{RQ3: What Monolingual Text Data Augmentation Techniques Have Been Used to Build Generative Language Models for Low-Resource Languages?}
    Given the prevalence and success of data augmentation in computer vision tasks, monolingual text augmentation techniques should be explored in further detail. Therefore, this section contains a discussion on the algorithms and processes presented in the results section for RQ3, that could be used to synthesise new training data for LRLs, in order of method frequency.

    \subsubsection{Paraphrasing}
    Paraphrasing techniques can be grouped into two primary categories: (1) word or sentence-level alterations, and (2) generation-powered paraphrasing.. 

    Word-level replacement strategies vary from study to study, however, they all share common risks. Much like the word-level grammatical checks used to improve the quality of generated QA pairs \cite{noauthor_qugan_nodate}, performing word-level augmentations by replacing terms with random or context-informed words could also distort the original semantic content of the sentence. The use of POS tag-based augmentations \cite{maimaiti_improving_2021} is an alternative process as choosing specific POS tags and replacing them could reduce the risk of grammatical errors produced through synthesising data. If a language has word classes that do not change based on agreement with other words in a text, a contextualised word replacement based on POS tags could produce more grammatically correct augmented sentences than random replacement. For example, adverbs tend not to affect agreement in English sentences, therefore they could be replaced with semantically similar terms without corrupting the grammar of the sentence. However, this approach is highly dependent on the quality of available POS tagging tools and models for a given LRL. Not only does the quality of POS-informed replacements hinge on the availability of POS tagging tools, the efficacy of the replacement strategy also directly impacts the quality of the paraphrased text. POS-based augmentations could reduce the risk of synthesising grammatically incorrect text, but the replacement strategy, neural-based or not, could corrupt or dilute the original meaning of the text being augmented. While this POS-informed approach is used to improve NMT for LRLs \cite{maimaiti_improving_2021}, this data augmentation method could also improve general language generation models, particularly for LRLs that consist of word classes which do not affect the grammar of other terms in a sentence. Future work could explore POS-informed replacements as a method for increasing language diversity across different generation tasks.
    
     Alternatively, generation-powered paraphrasing has shown limited results in generating new synthetic QA pairs \cite{noauthor_qugan_nodate}. However, the study's limited success in generating coherent text, even with BERT-powered grammar correction, is undermined by grammar characteristics being ignored in the target language. Although this approach of utilising a language model to paraphrase monolingual text data can produce grammatical errors, this could be compared to the distortions that can occur with back-translation. Therefore, it could be insightful to adopt paraphrasing tools to bolster text data in the context of machine translation tasks. In contrast, this potentially erroneous and nonsensical output could have a detrimental effect on the training of models for other generation tasks like dialogue generation and abstractive text summarisation where grammatical cohesion is expected. Although this generation method is prone to errors, LRL translation tasks could benefit from it as these errors are similar to the kind of grammatical errors that back-translation can produce.

\subsubsection{Grammatical Transformation}

This discussion section is split between providing a meta-analysis of sentence restructuring and introducing noise to augment monolingual data as forms of grammatical transformations.

The grammatical transformations category involves studies using a degree of sentence restructuring to produce synthetic sentence-level data. The use of a reverse transcription grammar model to generate diverse sentences for given input sentences can produce grammatically correct text \cite{wu_study_2023}, however, the reordering process could corrupt the original meaning of the sentence. Therefore, it is possible that this method could be beneficial for NMT tasks in the same way that back-translation can improve translation performance. In terms of generating text for dialogue generation or abstract summarisation, training text data containing many distortions of semantic content could produce poor models. However, this process of reordering a sentence could be beneficial for LRLs that have word order flexibility. Furthermore, this grammatical transformation method could be used to produce synthetic, parallel data for dialects that vary in word order from a standard form of a language. Future work could investigate the quality of synthetic text generated by reordering sentences to reflect dialect-specific syntax or word order. This would be beneficial as it could augment data for translation tasks between dialects.  

Alternatively, the success of applying noise to improve user-generated sequence-to-sequence translation performance \cite{marie_synthesizing_2020} should be explored for other generative modelling tasks. While this approach applies various stylistic changes such as introducing spelling errors, slang and informal expressions, the approach could be used for building grammar correction models. That is, this sequence-to-sequence approach could be used to fix grammatical agreement errors in sentences, where sentences with incorrect agreement could be mapped to sentences with correct agreement. Furthermore, datasets could be built for adapting or translating from one dialectal sentence structure to another within the same language. A dataset with a sentence in one dialect mapped to another could train a model for generating synthetic, parallel data between two language dialects. This would still follow the sequence-to-sequence nature of NMT but subsequently produce various augmentations of an original input sentence. This form of text data augmentation could effectively empower other generative tasks like dialogue generation and QA tasks to ensure grammatical quality and dialectal diversity for LRLs. 

\subsubsection{Data Enrichment}
While enriching existing data by appending additional information is a relatively straightforward approach, it remains underexplored in the literature \cite{usui_translation_2023}. Given that the impact of this data enrichment technique is not explored, more research is required to determine if adding metadata to text documents could help with text generation tasks. Including other data such as the access date, the year, the authors and the title of the training data could help to ground the generative language model through the inclusion of relevant information for each text example. However, this data enrichment approach does not introduce new semantic content; instead, it enhances the depth of the existing text data. Future work should empirically assess the effects of data enrichment in various generative tasks, particularly in LRL contexts where original content is scarce but supplementary metadata may be available.

    \subsection{RQ4: Which Sources Are Publishing in the Area of Low-Resource Language Modelling?}
    While the results section for RQ4 indicates a wide variety of sources publishing in the area of LRL modelling, the sources themselves serve different purposes. Given the dominance of ACL publications  spanning journals, workshops and conferences alike, it is clear that it has a central influence on the dissemination of literature relating to modelling LRLs. The high frequency of ACL publications from specific events such as the Empirical Methods in Natural Language Processing (EMNLP) and the Language Resources and Evaluation Conference (LREC) could suggest that these are key venues for LRL research. The prevalence of EMNLP papers could be attributed to the empirical nature of language modelling, and the frequency of LREC papers could be linked to the language resource element of building language models for LRLs. While ACM is associated with fewer events in comparison with other organisations such as ACL and IEEE, the Transactions on Asian and Low-Resource Language Information Processing journal is the most frequently sourced publication in this review, which could highlight that this journal is highly influential in terms of LRL modelling. 
    
    While arXiv preprint papers are not necessarily associated with traditional academic journals or undergo formal peer review, their prevalence could suggest that the field of LRL modelling is a rapidly evolving and current research topic. However, the frequency of arXiv papers could introduce limitations relating to the quality of research in the review process. We mitigate this potential bias by using the quality assessment questions relating to the rankings of conferences and journals. On the other hand, IEEE has more of a multi-disciplinary influence as it is the source of eight different publications relating to various fields within ICT and engineering. This diversity of publications could highlight that there are potential practical applications for LRL modelling in a variety of domains beyond purely theoretical research. Considering that Springer is the source of five journal papers in this review, this could indicate that there is a demand for topic-specific journals that relate to  tackling LRL modelling. Furthermore, individual contributions from other publishers such as Frontiers, the International Joint Conferences on Artificial Intelligence, and De Gruyter may suggest that the research area of LRL modelling is gaining broader recognition and rising in prevalence across publishers. Ultimately, the broad range of sources that span eight different publishers indicates that there is strong academic interest in the field of LRL modelling.

    \subsection{RQ5: What Architectures Are Being Used to Produce Low-Resource Language Models?}
    The prevalence of transformer-based models in this review reflects the transformative nature of the architecture for generative language modelling. Given the rise in popularity, adoption and performance of transformer-powered services such as ChatGPT, it is unsurprising that it dominates as the most frequently implemented model architecture. The sheer popularity of transformer-based models suggests that transformers are the most suitable architecture for modelling LRLs. The significant drop in popularity for RNNs, and more specifically LSTMs could reflect the evolution of the state-of-the-art architecture over time as RNN/LSTM-based approaches would have been the gold standard before the transformer was introduced \cite{mienye2024recurrent}. While relatively unpopular methods GAN, GRU and SMT-based methods have been used by studies in this review, their unpopularity could suggest that they are not the most optimal architectures for LRL generation in contrast with transformers. The limited use of these methods could reflect some attempts to use alternatives to the compute-heavy transformer architecture. For LRL communities that do not have access to sufficient computational power, such methods may still hold value as lightweight alternatives. 
    
    In terms of specific models, the popularity of GPT-2, the open-source predecessor to the widely popular GPT-3 and GPT-4 models by OpenAI, could indicate that it is suitable for LRL modelling. The relatively small size of GPT-2 small could make it appealing to LRL researchers, as it does not require as much training time and computational power to build it as other prevalent models such as the Llama3 family of models \cite{radford2019language, dubey2024llama}. The frequent use of XLM and sequence-to-sequence models such as mT5 and T5 could be attributed to the considerable number of studies in this review that tackle translation, an inherently sequence-to-sequence task. Ultimately, while there is a wide variety of models being used to build generative models for LRLs, transformer-based architectures dominate as the most frequently used for this task. Future work could focus on optimising transformer architectures for LRL modelling tasks. Given the potential lack of computational resources in LRL communities, investigating model efficiency improvements could lower the barrier to entry for research in the area. 

    \subsection{RQ6: How Effective Have Methods for Overcoming Data Scarcity Been at Improving Performance for Different NLG Tasks?}
    While some text data augmentation techniques are effective at improving the performance of different NLG models, it is worth considering how such methods compare across different NLG tasks. This review finds that translation as a generative task can be improved with several augmentation approaches such as back-translation \cite{pham_meta_2021,hong_cantonmt_2024,berckmann_low-resource_2020,yirmibesoglu_morphologically_2023,nissanka_exploring_2020,dandapat_iterative_2018,pham_van_improving_2022}, code-mixing \cite{bhowmick_improving_2023} and paraphrasing \cite{marie_synthesizing_2020, guo_automatically_2021}.  It is unlikely that back-translation would be an effective method for bolstering the amount of training data for building models for other LRL NLG tasks such as summarisation or dialogue generation.  Using back-translation for data synthesis could introduce semantic drift and inconsistencies at a syntactic level in the training corpus. Given that synthesising code mixed data has yielded performance increases for translation tasks, it could be insightful to apply this approach to languages that are often interwoven with each other. For example, some forms of modern Gaeilge usage could be more accurately modelled if it was trained on a corpus code mixed to include English, given the strong relationship between both languages \cite{laoire_irish-english_2016}. Therefore, it could be insightful to investigate the impact of code-mixing on other NLG tasks. While paraphrasing has produced performance improvements for translation tasks, it has also yielded improvements for other tasks such as question-answering corpus generation and dialogue generation \cite{noauthor_qugan_nodate,noauthor_paraphrase_nodate,wu_study_2023}. This approach of using paraphrasing models to bolster training data is a promising approach that hinges on the availability of enough data or existing paraphrasing models. More research and experimentation are required to assess the impact of large-scale paraphrasing as a means to augment training data for building NLG models for LRLs. Training multilingual and language family-based models in particular has been quite effective at bridging the data gap between LRLs. Given that this language family-based approach can reduce the number of model parameters \cite{cahyawijaya_indonlg_2021} and that multilingual modelling can be compared to a 33\% increase in training data for an LRL \cite{chang_when_2023}, it is clear that this is an effective text data augmentation method. It would be worth considering applying this multilingual, language family-based approach to various groups of related LRLs to further validate this observation. However, future work would be required to identify maximally related language family groupings and to collate existing resources across linguistic communities. While this multilingual, language family-based approach is promising from a general language modelling perspective, more research is required to assess  its impact on generative task-specific performance. 
    
    \subsection{RQ7: How Scarce Is the Data in Studies Tackling Low-Resource Language Modelling?}
    The availability of data for building language models differs considerably between LRLs. Therefore, it is important to highlight that measuring and comparing the scarcity of data presents a challenge given that the literature often reports different metrics relating to the training data. While some papers report the number of sentences used to train models for LRLs \cite{agarwal_zero-shot_2022,mao_tuning_2024,pham_van_improving_2022,hong_cantonmt_2024,gerz_language_2018}, others quantify by size \cite{noauthor_banglagpt_nodate,downey_targeted_2024,acikgoz_bridging_2024,cahyawijaya_indonlg_2021} or by tokens in their respective datasets \cite{toraman_llamaturk_2024, chang_when_2023,maimaiti_improving_2021}. On the other hand, some studies omit metrics relating to the size of their datasets completely \cite{niyogi_paramanu_2024,noauthor_english-arabic_nodate}. Overall, these varying methods of reporting dataset metrics present a limitation in comparing different language modelling approaches for LRLs, given that there is not a universal metric or framework used by these studies that allows us to evaluate and compare each dataset. Although it is a challenge to assess the scarcity across different studies, our analysis reveals two key insights. Firstly, the findings show that there is a gradient of data scarcity for LRLs. Specifically, some LRLs such as Khmer were modelled using three different datasets \cite{agarwal_zero-shot_2022,mao_tuning_2024,pham_van_improving_2022}, whereas many of the languages modelled were only trained on a single dataset \cite{berckmann_low-resource_2020, li_improving_2024}. While some LRLs such as Uyghur were modelled using a variety of datasets by different studies \cite{mao_tuning_2024,mi_multi-granularity_2024}, other LRLs have been modelled by a single study using much more data \cite{gerz_language_2018}. These findings suggest that there is a data disparity within the group of LRLs itself. 
    
    Secondly, out of the studies that quantify their data in terms of size, the Turkish dataset is significantly larger than the others \cite{acikgoz_bridging_2024}. This reflects the conflicting stance on the resource level of Turkish within the literature \cite{chang_when_2023,toraman_llamaturk_2024,yirmibesoglu_morphologically_2023} and consequently contributes to the need for a more robust framework for determining the resource level of a given language. Future work should focus on producing a universal framework or series of metrics for reporting on the resource level of a LRL. A universal framework would make comparisons between future LRL-based studies much easier. Additionally, it would enable future researchers to find the most underrepresented LRLs and gauge their suitability for generative modelling based on the available data for a given language. Future work on systematically reporting data resources could contribute to a more comprehensive understanding of cross-lingual resources for modelling multiple languages or language families. Overall, this review finds that the lack of standardisation for reporting available data, combined with the existing data disparity among LRLs, presents a barrier to progress in generative LRL modelling.    
    
    \subsection{RQ8: How Are the Models in These Studies Being Evaluated?}
    These studies evaluate generative language models for LRLs in many different ways due to the broad range of NLG tasks being modelled. Similarly to the challenges of quantifying dataset size in RQ7, the diversity of reported evaluation metrics introduces challenges when comparing results between studies. Given that half of the papers in this review tackle translation, it is unsurprising that BLEU dominates as the most frequently used evaluation metric. However, it is worth highlighting that the BLEU metric suffers from several known shortcomings relating to the reporting of results. \textit{Post} highlights that preprocessing techniques can have a significant impact on BLEU scores, thus impeding  comparability between studies \cite{post_call_2018}. The use of sacreBLEU as a means to address BLEU-based limitations \cite{wongso_many--many_2023, hong_cantonmt_2024,liu_low-resource_2022} and metrics such as ROUGE, chrF, METEOR and relatively newer methods like COMET and sacreBLEU indicates that a wide variety of fine-tuned task-specific models are being trained. 
    
    However, the infrequent use of human evaluation methods reveals considerable limitations in most of the studies. Given that modelling an LRL contributes to cultural preservation, evaluating whether a model successfully captures the qualities of a language should be essential in determining if an LRL has been modelled accurately. Therefore, in the interest of ethical language modelling, studies should not prematurely report success when modelling a LRL without a degree of human evaluation, as the statistical metrics do not systematically quantify whether a model for a LRL is representative of it or not. The lack of a universal approach to evaluating LRL models and the use of flawed and dated, though still prevalent, metrics presents a challenge when comparing the results of different studies within this field of generative language modelling. Future research should focus on building a framework for evaluating LRL models across the spectrum of generative tasks. Additionally, future work should endeavour to include human, qualitative evaluation when evaluating a model's capacity to generate text in a given LRL. 
    
    \subsection{RQ9: What Is the NLG Task?}
    Given that this review focuses on modelling LRLs, the prevalence of translation may be attributed to the need for translation models as an initial step in digitising and creating other generative models for these languages. Furthermore, the resources for building translation models are likely more readily available in contrast with the extensive resources that are required for training other generative models such as dialogue generation models. The predominance of translation suggests that LRLs lag significantly behind widely spoken languages such as English, both in research attention and in the availability of supporting resources. On the other hand, studies contributing a general language model that can be adapted and fine-tuned for different tasks \cite{gerz_language_2018, noauthor_banglagpt_nodate} could be an initial step that would yield further research relating to downstream generative language tasks. 
    
    Additionally, the prevalence of question-answering models likely reflects a demand for information retrieval systems for LRLs. Given that other NLG tasks such as summarisation and dialogue generation are rarely modelled, it is clear that LRLs are lacking in research for downstream beyond translation and QA tasks. This is likely due to the lack of large, complex and task-specific labelled corpora required to train such downstream models. Furthermore, a pre-trained model is typically required as a base for fine-tuning task-specific models. Therefore, future research should be focused on releasing open-source pre-trained models for LRLs, as this would enable researchers to tackle other downstream generative tasks. Given the current popularity of generative language models, it is likely that the frequency of NLG tasks will shift from translation to general language model research and downstream generative tasks in the future. 
    
    However, it is worth considering that, in theory, a strong translation model could be the key to unlocking generative technology access for LRL communities. More specifically, an effective translation model could interface with the input and output of a better-resourced, higher-performing model for majority languages, such as English. This could be a sustainable way of reducing language inequality in the digital world as such translation models could wrap around future state-of-the-art models with relative ease. Future work should investigate the feasibility and quality of output when wrapping existing generative models with a LRL translation model.   

\section{Limitations} 
This section outlines the considerations and limitations relating to this systematic review.
Given that LRLs are a central element of this research, excluding non-English papers could omit relevant literature that addresses the challenge of building generative language models with limited data. Furthermore, the title, abstract and keywords of a paper were examined to identify their relevance during the second pass of the study selection phase. This introduces a limitation in this systematic review as relevant information could have been present in the identified papers but absent from their respective metadata. Although both popular platforms and NLP-specific repositories were searched for relevant studies, relevant studies could exist in less conventional, but still significant sources such as niche linguistic journals or conferences. While the inclusion of synonyms for keywords such as "low resource language" and "language model" improves coverage, it does not guarantee that all relevant studies will be identified. In terms of performing the study selection, the use of ASReview as a tool to improve the initial screening process could introduce inaccuracies. The tool does not classify studies but instead predicts their relevance using active learning. Therefore, algorithmic bias could be introduced whereby a reviewer could expect studies seen at the beginning of the selection process to be relevant and could consequently include irrelevant studies. It is worth reporting that a single reviewer worked independently on each stage of the screening process. Although this increases the risk of subjectivity and errors, this reflects the challenges involved with sourcing adequate resources such as multiple reviewers to mitigate bias during the study selection phase. The heterogeneity of results across the different studies imposes a limit on the validity of the comparisons and conclusions between them. Many of the studies report the same metrics such as BLEU, however not all of them include a detailed description of their experimental setup. Moreover, some studies report different or updated forms of metrics (sacreBLEU) for the same generative tasks, making accurate comparisons of their results challenging. 

\section{Conclusion}
This systematic review has addressed the challenges posed by data scarcity when building generative language models for LRLs. We have extracted relevant information that contributes to the overall research question by clearly outlining objectives relating to identifying methods for overcoming data scarcity, their impact on model performance and documenting future research directions. By applying a screening process and a series of sub-research questions to each paper identified in the study selection a collection of results was produced.

The results of this systematic review indicate that Indo-European languages dominate LRL modelling research, suggesting that there are disparities within the category of LRLs itself. These findings could suggest that there is a need to define a more nuanced measure of LRL data scarcity. A scale could help to inform a suitable method for overcoming data scarcity. The results also reveal monolingual text data augmentation, multilingual modelling and back-translation were the most commonly used augmentation techniques. Monolingual text data augmentations were found to be effective at improving LRL models, however its efficacy hinges on both the availability and quality of existing language models for neural-based replacement approaches. Therefore, further research is needed to develop NLP models for underrepresented LRLs that currently lack sufficient models, enabling them to benefit from the augmentations discussed earlier. In terms of model architectures, the results indicate that transformer-based models dominate across LRL modelling research. Their popularity can be attributed to their success in modelling a wide range of languages. Although BLEU was the most common evaluation metric, there is a need for universal evaluation metrics that can be used to compare the performance across different LRLs. Universal approaches for reporting the quantity of data used to train LRL models need to be adopted, as this systematic review found that inconsistent reporting made it challenging to compare data scarcity across different studies. Publications in the field of LRL modelling are predominantly driven by ACL and ACM, with a significant number of preprints on arXiv potentially indicating that the field is evolving. The results indicate that translation dominated as the most common NLG task. This could suggest that LRL research prioritises foundational language applications like machine translation. 

More research needs to be done to bridge the gap in language inequality and to ensure that less-resourced languages, and communities by extension, are not left behind in the digital age. Developing and releasing conversational language models akin to ChatGPT for LRLs could be a significant step in this direction and consequently requires further efforts. Not only do these LRL models have the potential to preserve entire languages by capturing their linguistic features, but they are also essential for sustaining and promoting language diversity in an age increasingly dominated by large-scale language technologies. The findings of this systematic review aim to support researchers working with low-resource languages by providing a clearer understanding of current strategies for overcoming data scarcity across diverse NLG tasks. Ultimately, this knowledge can contribute to empowering speakers of minority and underrepresented languages through more inclusive technological development.

\section{Statements and Declarations}
\textbf{Conflict of Interest} The authors declare that they have no conflicts of interest related to this work.

\section{Acknowledgments}
This publication has emanated from research conducted with the financial support of Taighde Éireann - Research Ireland under Grant number 18/CRT/6223.

\bibliographystyle{ACM-Reference-Format}
\bibliography{references}

\section{Appendix A}
See Table~\ref{tab:newrq12},~\ref{tab:newrq45} and~\ref{tab:newrq789}.
\begin{table*}[t]
   \caption{Data extracted for RQ1, RQ2, and RQ3. The RQ1 column lists the low-resource languages addressed by each study. RQ2 presents the technical methods used to overcome data scarcity. RQ3 identifies the text data augmentation techniques used.}

    \label{tab:newrq12}
    \centering
    {\tiny  
        \begin{tabular}{p{0.25cm} p{10cm} p{3cm} p{3cm}} 
            \toprule
            Paper & RQ1 & RQ2 & RQ3 \\
            \midrule
            \cite{li_improving_2024} & Kashmiri & Prompt engineering & -\\
\cite{noauthor_qugan_nodate} & Tibetan & Augment monolingual data, monolingual model & Paraphrase\\
\cite{zhang_chren_2020} & Cherokee & Back-translation & -\\
\cite{usui_translation_2023} & Historical Japanese & Augment monolingual data & Data enrichment\\
\cite{sorokin_ask_2022} & Arabian, Bengali, Telugu & Prompt engineering & -\\
\cite{gao_soft_2019} & - & Augment monolingual data & Soft Contextual Data Augmentation, paraphrase\\
\cite{bhowmick_improving_2023} & Hindi, Bengali & Code mixing, multilingual models & -\\
\cite{otegi_conversational_2020} & Basque & Adaptive learning & -\\
\cite{marie_synthesizing_2020} & - & Back-translation, augment monolingual data & Re-ordering monolingual sentences, paraphrase\\
\cite{ahmadnia_augmenting_2019} & Persian & Back-translation & -\\
\cite{yirmibesoglu_morphologically_2023} & Turkish & Back-translation & -\\
\cite{maimaiti_improving_2021} & Azerbaijani, Hindi & Augment monolingual data & Replace main POS tags, paraphrase\\
\cite{cahyawijaya_indonlg_2021} & Sudanese, Javanese & Family of LRLs, multilingual models & -\\
\cite{guo_automatically_2021} & - & Augment monolingual data & Paraphrase\\
\cite{wongso_many--many_2023} & Balinese, TobaBatak, BatakSimalungun, BatakKaro, Buginese, Javanese, Madurese, Makassarese, TorajaSadan, Sundanese, Ambonese, Acehnese, AralleTabulahan, Berik, Balantak, PakpakDairi, Bauzi, Galela, Gorontalo, Hawu, Iban, Abun, DaaKaili, LampungApi, Meyah, Minangkabau, Kupang, Mongondow, Mamas, Manggarai, Duri, Mentawai, Nias, Ngaju, Napu, Pamona, Uma, Bambam, Sasak, Sangir, Tabaru, Termanu, Anggurk, Yawa & Family of LRLs, multilingual models & -\\
\cite{dandapat_iterative_2018} & Telugu & Back-translation & -\\
\cite{wang_switchout_2018} & - & Back-translation, augment monolingual data & -\\
\cite{noauthor_paraphrase_nodate} & - & Augment monolingual data & Paraphrase\\
\cite{pham_meta_2021} & Azerbaijani, Belarusian, Glacian, Slovak & Back-translation & -\\
\cite{niyogi_paramanu_2024} & Assamese, Bengali, Hindi, Konkani, Maithili, Marathi, Odia, Sanskrit, Tamil, Telugu & Multilingual models, family of LRLs & -\\
\cite{downey_targeted_2024} & Estonian, Komi, Mari, Erzya, Veps, Udmurt, Sámi, Karelian, Moksha, Livonian, Votic, Ingrian & Multilingual models, family of LRLs & -\\
\cite{hong_cantonmt_2024} & Cantonese & Back-translation & -\\
\cite{chang_when_2023} & Turkish, Lithuanian, Hindi, Catalan, Slovak, NorwegianBokmal, Estonian, Bengali, Latvian, Serbian, Slovenian, Tamil, Albanian, Azerbaijani, Urdu, Nepali, Macedonian, Kazakh, Georgian, Armenian, Belarusian, Esperanto, Croatian, Malayalam, Icelandic, Welsh, Telugu, Galician, Hausa, Mongolian, Marathi, Asturian, Afrikaans, Basque, Burmese, Bosnian, CentralKanuri, Somali, Tatar, Cebuano, Kannada, CentralKhmer, Gujarati, Panjabi, Bashkir, CentralKurdish, Maltese, Serbo, Tajik, Tagalog, Kirghiz, Tigrinya, Malay, Igbo, Sinhala, Irish, Amharic, Uzbek, Swahili, Luxembourgish, Yoruba, Haitian, Kinyarwanda, Samoan, Javanese, NorwegianNynorsk, Lao, Nyanja, Sindhi, SouthernPashto, Sundanese, Maori, Occitan, PlateauMalagasy, Pushto, ScottishGaelic, Shona, Waray, Zulu, Dari, NorthernUzbek, Uighur, Assamese, SouthernSotho, Lushai, StandardMalay, Xhosa, Sicilian, Lombard, EasternYiddish, EgyptianArabic, Limburgan, Odia, SouthAzerbaijani, AyacuchoQuechua, WestCentralOromo, HalhMongolian, Venetian, Banjar, Gilaki, Ganda, Papiamento, Sanskrit, Rundi, Achinese, Tswana, WesternPanjabi, Twi, Iloko, Chechen, Tsonga, Yakut, WesternFrisian, Kurdish, Ewe, Oriya, Latin, Chuvash, Minangkabau, Faroese, Breton, YueChinese, Pedi, ToskAlbanian, CrimeanTatar, NorthernKurdish, Kabyle, Fon, LowGerman, Inuktitut, Maithili, Lingala, Guarani, Tibetan, Pangasinan, Bemba, Wolof, Tumbuka, Luo, Malagasy, Oromo, Dimli, Yiddish, Tuvinian, MinNanChinese, Balinese, Fijian, CentralAymara, Aragonese, Ligurian, Dhivehi, Luba, Silesian, NigerianFulfulde, SwissGerman, Swati, Betawi, Friulian, Sardinian, Bavarian, TokPisin, Umbundu, NigerianPidgin, EasternMari, Ido, RussiaBuriat, Bhojpuri, Bambara, Chokwe, SouthwesternDinka, Dyula, Mossi, Turkmen, Piemontese, WuChinese, Kongo, Dargwa, Buginese, Kabuverdianu, Kabiye, Kimbundu, Hawaiian, Sango, Mirandese, Kachin, Ingush, Kikuyu, Romansh, Kaqchikel, Kabardian, Volapuk, MandarinChinese, Kituba, Magahi, CentralBikol, Kashmiri, CuscoQuechua, LiteraryChinese, Walloon, Akan, Berber, Chhattisgarhi, Interlingua, UpperSorbian, Latgalian, Santali, Susu, Nuer, Vlaams, Quechua, Udmurt, Veps, Avaric, Swahili, Lak, Erzya, Urdu, Ossetian, Uighur, Lezghian, GoanKonkani, Shan, Serbian & Multilingual models & -\\
\cite{shafayat_benqa_2024} & Bengali & Mass translation, augment monolingual data & Mass translation\\
\cite{tanwar_translating_2020} & Telugu, Tamil, Gujarati, Punjabi, Hindi & Multilingual models & -\\
\cite{gerz_language_2018} & Amharic, Catalan, Greek, Estonian, Basque, Farsi, Hindi, Croatian, Javanese, Georgian, Khmer, Kannada, Lithuanian, Latvian, Malay, Mongolian, Burmese, MinNan, Norwegian, Slovak, Slovene, Serbian, Tamil, Tagalog, Turkish & Multilingual models & -\\
\cite{berckmann_low-resource_2020} & UpperSorbian & Back-translation & -\\
\cite{liu_low-resource_2022} & Cantonese & Multilingual models & -\\
\cite{pham_van_improving_2022} & Khmer & Back-translation & -\\
\cite{nissanka_exploring_2020} & Tamil, Sinhala & Back-translation & -\\
\cite{agarwal_zero-shot_2022} & Bengali, Telegu, Khmer, Malay & Prompt engineering & -\\
\cite{scalvini_evaluating_2024} & Faroese & Prompt engineering & -\\
\cite{guo_teaching_2024} & - & Prompt engineering & -\\
\cite{noauthor_frontiers_nodate} & Slovenian & Monolingual model & -\\
\cite{bendel_llegra_2024} & Vallader & Prompt engineering & -\\
\cite{wu_study_2023} & - & Augment monolingual data & Re-ordering monolingual sentences\\
\cite{babaali_breaking_2024} & Algerian & Multilingual models & -\\
\cite{li_multi-tasking_2022} & Tibetan, Mongolian, Uyghur & Multilingual models, family of LRLs & -\\
\cite{wang_survey_2022} & - & Back-translation & -\\
\cite{mi_multi-granularity_2024} & Uyghur & Monolingual model & -\\
\cite{noauthor_efficient_nodate} & - & Adaptive learning & -\\
\cite{wang2021survey} & - & Augment monolingual data & Fadaee, re-ordering monolingual sentences\\
\cite{noauthor_english-arabic_nodate} & - & Adaptive learning & -\\
\cite{mao_tuning_2024} & Afrikaans, Amharic, Belarusian, Welsh, Irish, Scottish, Galician, Hausa, Georgian, Kazakh, Khmer, Kyrgyz, Limburgish, Myanmar, Bokmål, Nynorsk, Occitan, Sinhala, Tajik, Turkmen, Tatar, Uighur, Uzbek, Yiddish & Prompt engineering & -\\
\cite{toraman_llamaturk_2024} & Turkish & Adaptive learning, prompt engineering, vocab extension & -\\
\cite{zhang_teaching_2024} & Zhuang & Prompt engineering & -\\
\cite{acikgoz_bridging_2024} & Turkish & Adaptive learning, monolingual model & -\\
\cite{haque_recent_2021} & - & Mass translation, augment monolingual data & Mass translation, Fadaee, Soft Contextual Data Augmentation, SwitchOut\\
\cite{shi_low-resource_2022} & - & Back-translation, augment monolingual data & Fadaee, SwitchOut, SMRT Simulated multiple reference training, replace main POS tags \\
\cite{zhang_neural_2024} & - & Augment monolingual data & Re-ordering monolingual sentences, paraphrase\\
\cite{noauthor_banglagpt_nodate} & Bengali & Monolingual model & -\\
\cite{noauthor_bidirectional_nodate} & Marathi & Prompt engineering & -\\
\cite{noauthor_qasina_nodate} & Indonesian & Monolingual model, adaptive learning & -\\
\cite{noauthor_generative_nodate} & Odia & Prompt engineering & -\\
            \bottomrule
        \end{tabular}
    }
\end{table*}
\begin{table*}[t]
    \caption{Data extracted for RQ4 and RQ5. The RQ4 column identifies the publisher of each study. The publication/venue column indicates the source of each paper. The location column refers to the location of the research institution affiliated with the first author of each paper. The date column denotes the year of publication for each study. The RQ5 architecture column describes the general model architecture used by each paper. The RQ5 model column lists specific models employed by each paper. A hyphen (-) indicates the absence of an applicable field.     } 
    \label{tab:newrq45}
    \centering
    {\tiny  
        \begin{tabular}{p{0.5cm} p{1cm} p{6cm} p{1cm} p{0.5cm} p{3cm} p{3cm}} 
            \toprule
            Paper & RQ4 & Publication/Venue & Location & Date & RQ5 Architecture & RQ5 Model \\
            \midrule
            \cite{li_improving_2024} & ACL & ACL EACL - The European Chapter of the ACL & USA & 2024 & Transformer & mt5, flant5base \\
\cite{noauthor_qugan_nodate} & IEEE & IEEE Access & China & 2019 & RNN, Transformer, GAN & QuGAN, BERT \\
\cite{zhang_chren_2020} & ACL & ACL EMNLP - Empirical Methods in Natural Language Processing & USA & 2020 & RNN, Transformer & BERT \\
\cite{usui_translation_2023} & ACL & ACL NLP4DH - International Conference on Natural Language Processing for Digital Humanities & Japan & 2023 & Transformer & t5 \\
\cite{sorokin_ask_2022} & ACL & ACL NAACL - North American Chapter of the Association for Computational Linguistics & Russia & 2022 & Transformer & XLM, RoBERTa \\
\cite{gao_soft_2019} & ACL & ACL - Annual Meeting of the Association for Computational Linguistics & China & 2019 & Transformer & - \\
\cite{bhowmick_improving_2023} & ACM & ACM TALLIP - Transactions on Asian and Low-Resource Language Information Processing & India & 2023 & Transformer & mT5 \\
\cite{otegi_conversational_2020} & ACL & ACL LREC - International Conference on Language Resources and Evaluation & Spain & 2020 & Transformer & mBERT, BERTeus \\
\cite{marie_synthesizing_2020} & ACL & ACL - Transactions of the Association for Computational Linguistics & Japan & 2020 & Transformer & XLM \\
\cite{ahmadnia_augmenting_2019} & De Gruyter & De Gruyter - Open Computer Science & USA & 2019 & LSTM, RNN & - \\
\cite{yirmibesoglu_morphologically_2023} & ACM & ACM TALLIP - Transactions on Asian and Low-Resource Language Information Processing & Turkey & 2023 & Transformer, BiLSTM, RNN, LSTM & - \\
\cite{maimaiti_improving_2021} & ACM & ACM TALLIP - Transactions on Asian and Low-Resource Language Information Processing & China & 2021 & Transformer & - \\
\cite{cahyawijaya_indonlg_2021} & ACL & ACL EMNLP - Empirical Methods in Natural Language Processing & Hong Kong & 2021 & Transformer & mBART, GPT2 \\
\cite{guo_automatically_2021} & IJCAI & IJCAI - International Joint Conference on Artificial Intelligence & China & 2021 & set2sequence & - \\
\cite{wongso_many--many_2023} & IEEE & IEEE Access & Indonesia & 2023 & Transformer & mt5 \\
\cite{dandapat_iterative_2018} & ACL & ACL - Proceedings of the 21st Annual Conference of the European Association for Machine Translation & USA & 2018 & SMT, Transformer & mBERT \\
\cite{wang_switchout_2018} & ACL & ACL EMNLP - Empirical Methods in Natural Language Processing & USA & 2018 & Transformer & - \\
\cite{noauthor_paraphrase_nodate} & ACL & ACL - Annual Meeting of the Association for Computational Linguistics & China & 2020 & Seq2Seq, Transformer & - \\
\cite{pham_meta_2021} & ARXIV & ARXIV & USA & 2021 & Transformer & - \\
\cite{niyogi_paramanu_2024} & ARXIV & ARXIV & India & 2024 & Transformer & - \\
\cite{downey_targeted_2024} & ARXIV & ARXIV & USA & 2024 & Transformer & XLM \\
\cite{hong_cantonmt_2024} & ARXIV & ARXIV & UK & 2024 & Transformer & mBART \\
\cite{chang_when_2023} & ARXIV & ARXIV & USA & 2023 & Transformer & GPT2 \\
\cite{shafayat_benqa_2024} & ARXIV & ARXIV & Korea & 2024 & Transformer & CLaude2.1, LLama27b, LLama213b, Mistral7b \\
\cite{tanwar_translating_2020} & ACM & ACM TALLIP - Transactions on Asian and Low-Resource Language Information Processing & India & 2020 & GAN & - \\
\cite{gerz_language_2018} & ACL & ACL - Transactions of the Association for Computational Linguistics & UK & 2018 & LSTM, RNN & - \\
\cite{berckmann_low-resource_2020} & ACL & ACL WMT - Conference on Machine Translation & USA & 2020 & Transformer & GPT2 \\
\cite{liu_low-resource_2022} & ACL & ACL - Workshop on NLP for Similar Languages, Varieties and Dialects & Sweden & 2022 & Transformer, BiLSTM, RNN, LSTM & - \\
\cite{pham_van_improving_2022} & ACM & ACM SoICT - Symposium on Information and Communication Technology & Vietnam & 2022 & Transformer & mBART50 \\
\cite{nissanka_exploring_2020} & IEEE & IEEE - International Conference on Advances in ICT for Emerging Regions & Sri Lanka & 2020 & LSTM, RNN & - \\
\cite{agarwal_zero-shot_2022} & ACL & ACL - Proceedings of the Workshop on Multilingual Information Access (MIA) & USA & 2022 & Transformer & mt5, mBERT, CORA \\
\cite{scalvini_evaluating_2024} & ACL & ACL LREC - International Conference on Language Resources and Evaluation & Faroe Islands & 2024 & Transformer & - \\
\cite{guo_teaching_2024} & ACL & ACL LREC - Joint International Conference on Computational Linguistics, Language Resources and Evaluation & China & 2024 & Transformer & - \\
\cite{noauthor_frontiers_nodate} & Frontiers & Frontiers in Artificial Intelligence & Slovenia & 2023 & Transformer & t5 \\
\cite{bendel_llegra_2024} & Springer & Springer - International Journal of Information Technology & Switzerland & 2024 & Transformer & - \\
\cite{wu_study_2023} & ACM & ACM APIT - Asia Pacific Information Technology Conference & China & 2023 & - & - \\
\cite{babaali_breaking_2024} & Springer & Springer - International Journal of Information Technology & Algeria & 2024 & Transformer & - \\
\cite{li_multi-tasking_2022} & Springer & Springer - Machine Translation & China & 2022 & Transformer & - \\
\cite{wang_survey_2022} & ACL & ACL - Transactions of the Association for Computational Linguistics & China & 2022 & - & - \\
\cite{mi_multi-granularity_2024} & ACM & ACM TALLIP - Transactions on Asian and Low-Resource Language Information Processing & China & 2024 & Transformer & - \\
\cite{noauthor_efficient_nodate} & ACM & ACM TALLIP - Transactions on Asian and Low-Resource Language Information Processing & China & 2020 & Transformer & - \\
\cite{wang2021survey} & ARXIV & ARXIV & China & 2021 & - & - \\
\cite{noauthor_english-arabic_nodate} & IEEE & ACS/IEEE - International Conference on Computer Systems and Applications & Egypt & 2023 & Transformer, LSTM, RNN & Helskini \\
\cite{mao_tuning_2024} & ACL & ACL LoResMT - Workshop on Technologies for Machine Translation of Low-Resource Languages & USA & 2024 & Transformer & Bloomz \\
\cite{toraman_llamaturk_2024} & ARXIV & ARXIV & Turkey & 2024 & Transformer & LLaMa7b \\
\cite{zhang_teaching_2024} & ARXIV & ARXIV & China & 2024 & - & - \\
\cite{acikgoz_bridging_2024} & ARXIV & ARXIV & Turkey & 2024 & Transformer & Mistral7B, GPT2 \\
\cite{haque_recent_2021} & Springer & Springer - Machine Translation & Ireland & 2021 & - & - \\
\cite{shi_low-resource_2022} & ACM & ACM TALLIP - Transactions on Asian and Low-Resource Language Information Processing & China & 2022 & - & - \\
\cite{zhang_neural_2024} & ACM & ACM TALLIP - Transactions on Asian and Low-Resource Language Information Processing & Japan & 2024 & - & - \\
\cite{noauthor_banglagpt_nodate} & IEEE & IEEE ICICT4SD - International Conference on Information and Communication Technology for Sustainable Development & Bangladesh & 2023 & Transformer & GPT2 \\
\cite{noauthor_bidirectional_nodate} & IEEE & IEEE INCOFT - International Conference on Futuristic Technologies & India & 2023 & - & - \\
\cite{noauthor_qasina_nodate} & IEEE & IEEE ICAICTA - International Conference of Advanced Informatics: Concept, Theory and Application & Indonesia & 2023 & Transformer & mBERT, XLM, IndoBERT \\
\cite{noauthor_generative_nodate} & IEEE & IEEE CCPIS - International Conference on Circuits, Power and Intelligent Systems & India & 2023 & Transformer & - \\
            \bottomrule
        \end{tabular}
    }
\end{table*}
\begin{table*}[t]
    \caption{Data extracted for RQ7, RQ8, and RQ9. The RQ7 column describes the scarcity of data available for low-resource languages in each study, in terms of sentences (S), tokens (T), and size (GB), where k, m, and b represent thousands, millions, and billions, respectively. The RQ8 column lists the evaluation metrics and the RQ9 column categorises the NLG tasks addressed by each study.}
    \label{tab:newrq789}
    \centering
    {\tiny  
        \begin{tabular}{p{0.25cm} p{10cm} p{3cm} p{3cm}} 
            \toprule
            Paper & RQ7 & RQ8 & RQ9 \\
            \midrule
            \cite{li_improving_2024} & Kashmiri: 26kS & BLEU, BERTScore, chrf++ & Translation \\
\cite{noauthor_qugan_nodate} & Tibetan: 22kS & BLEU & Question Answering \\
\cite{zhang_chren_2020} & Cherokee: 14kS & BLEU & Translation \\
\cite{usui_translation_2023} & - & BLEU & Translation \\
\cite{sorokin_ask_2022} & - & Recall, F1-score, exact match & Question Answering \\
\cite{gao_soft_2019} & - & BLEU & Translation \\
\cite{bhowmick_improving_2023} & Hindi: 5kS, Bengali: 5kS & BLEU, ROUGE, METEOR & Translation \\
\cite{otegi_conversational_2020} & Basque: 2kS & F1-score & Question Answering \\
\cite{marie_synthesizing_2020} & - & BLEU & Translation \\
\cite{ahmadnia_augmenting_2019} & - & BLEU & Translation \\
\cite{yirmibesoglu_morphologically_2023} & Turkish: 207kS & BLEU & Translation \\
\cite{maimaiti_improving_2021} & Azerbaijan: 3.1mT, Hindi: 7.3mT, Uzbek: 2.2mT, Turkish: 0.8mT & BLEU, METEOR & Translation \\
\cite{cahyawijaya_indonlg_2021} & - & BLEU, Xtreme Indosum, TyDiQA, XPersona & Language model, translation, summarisation, conversation, question answering \\
\cite{guo_automatically_2021} & - & BLEU, ROUGE & Translation \\
\cite{wongso_many--many_2023} & - & SacreBLEU & Translation \\
\cite{dandapat_iterative_2018} & Telugu: 750kS & BLEU, human evaluation & Translation \\
\cite{wang_switchout_2018} & - & BLEU & Translation \\
\cite{noauthor_paraphrase_nodate} & - & BLEU, Entity match rate, Success F1 & Conversation \\
\cite{pham_meta_2021} & Azerbaijani: 5946S,  Glacian: 10kS & BLEU & Translation \\
\cite{niyogi_paramanu_2024} & - & Human evaluation & Language model \\
\cite{downey_targeted_2024} & Estonian: 6.1GB & Accuracy, unlabeled attachment score & Language model \\
\cite{hong_cantonmt_2024} & Cantonese: 1.1mS & SacreBLEU, hLEPOR, COMET, BERTscore, Human evaluation & Translation \\
\cite{chang_when_2023} & Iranian Persian: 1bT, Modern Greek: 1bT, Standard Arabic: 1bT, Turkish: 1bT, Lithuanian: 100mT, Hindi: 100mT, Catalan: 100mT, Slovak: 100mT, Norwegian Bokmal: 100mT, Estonian: 100mT, Bengali: 100mT, Latvian: 100mT, Serbian: 100mT, Slovenian: 100mT, Tamil: 100mT, Albanian: 100mT, Azerbaijani: 100mT, Urdu: 100mT, Nepali: 100mT, Macedonian: 100mT, Kazakh: 100mT, Georgian: 100mT, Armenian: 100mT, Belarusian: 100mT, Esperanto: 10mT, Croatian: 10mT, Malayalam: 10mT, Icelandic: 10mT, Welsh: 10mT, Telugu: 10mT, Galician: 10mT, Hausa: 10mT, Mongolian: 10mT, Marathi: 10mT, Asturian: 10mT, Afrikaans: 10mT, Basque: 10mT, Burmese: 10mT, Bosnian: 10mT, Central Kanuri: 10mT, Somali: 10mT, Tatar: 10mT, Cebuano: 10mT, Kannada: 10mT, Central Khmer: 10mT, Gujarati: 10mT, Panjabi: 10mT, Bashkir: 10mT, Central Kurdish: 10mT, Maltese: 10mT, Serbo-Croatian: 10mT, Tajik: 10mT, Tagalog: 10mT, Kirghiz: 10mT, Tigrinya: 10mT, Malay: 10mT, Igbo: 10mT, Sinhala: 10mT, Irish: 10mT, Amharic: 10mT, Uzbek: 10mT, Swahili: 10mT, Luxembourgish: 10mT, Yoruba: 10mT, Haitian: 10mT, Kinyarwanda: 10mT, Samoan: 10mT, Javanese: 10mT, Norwegian Nynorsk: 10mT, Lao: 10mT, Nyanja: 10mT, Sindhi: 10mT, Southern Pashto: 10mT, Sundanese: 10mT, Maori: 10mT, Occitan: 10mT, Plateau Malagasy: 10mT, Pushto: 10mT, Scottish Gaelic: 10mT, Shona: 10mT, Waray: 10mT, Zulu: 10mT, Dari: 10mT, Northern Uzbek: 10mT, Uighur: 10mT, Assamese: 10mT, Southern Sotho: 10mT, Lushai: 1mT, Standard Malay: 1mT, Xhosa: 1mT, Sicilian: 1mT, Lombard: 1mT, Eastern Yiddish: 1mT, Egyptian Arabic: 1mT, Limburgan: 1mT, Odia: 1mT, South Azerbaijani: 1mT, Ayacucho Quechua: 1mT, West Central Oromo: 1mT, Halh Mongolian: 1mT, Venetian: 1mT, Banjar: 1mT, Gilaki: 1mT, Ganda: 1mT, Papiamento: 1mT, Sanskrit: 1mT, Rundi: 1mT, Chinese Hant: 1mT, Achinese: 1mT, Tswana: 1mT, Western Panjabi: 1mT, Twi: 1mT, Iloko: 1mT, Chechen: 1mT, Tsonga: 1mT, Yakut: 1mT, Western Frisian: 1mT, Kurdish: 1mT, Ewe: 1mT, Oriya: 1mT, Latin: 1mT, Chuvash: 1mT, Minangkabau: 1mT, Faroese: 1mT, Breton: 1mT, Yue Chinese: 1mT, Pedi: 1mT, Tosk Albanian: 1mT, Crimean Tatar: 1mT, Northern Kurdish: 1mT, Kabyle: 1mT, Fon: 1mT, Low German: 1mT, Inuktitut: 1mT, Maithili: 1mT, Lingala: 1mT, Guarani: 1mT, Tibetan: 1mT, Pangasinan: 1mT, Bemba: 1mT, Wolof: 1mT, Tumbuka: 1mT, Luo: 1mT, Malagasy: 1mT, Oromo: 1mT, Dimli: 1mT, Yiddish: 1mT, Tuvinian: 1mT, Min Nan Chinese: 1mT, Balinese: 1mT, Fijian: 1mT, Central Aymara: 1mT, Aragonese: 1mT, Ligurian: 1mT, Dhivehi: 1mT, Luba-Lulua: 1mT, Silesian: 1mT, Nigerian Fulfulde: 1mT, Swiss German: 1mT, Swati: 1mT, Betawi: 1mT, Friulian: 1mT, Sardinian: 1mT, Bavarian: 1mT, Tok Pisin: 1mT, Umbundu: 1mT, Nigerian Pidgin: 1mT, Eastern Mari: 1mT, Ido: 1mT, Russia Buriat: 1mT, Bhojpuri: 1mT, Bambara: 1mT, Chokwe: 1mT, Southwestern Dinka: 1mT, Dyula: 1mT, Mossi: 1mT, Turkmen: 1mT, Piemontese: 1mT, Central Kanuri: 1mT, Wu Chinese: 1mT, Kongo: 1mT, Dargwa: 1mT, Buginese: 1mT & Perplexity, log-likelihood & Language model \\
\cite{shafayat_benqa_2024} & Bengali: 5kS & COPA & Question Answering \\
\cite{tanwar_translating_2020} & Telugu: 70kS,Tamil: 70kS,Gujarati: 70kS,Punjabi: 70kS,Hindi: 70kS & BLEU, TER & Translation \\
\cite{gerz_language_2018} & Amharic: 511kS, Catalan: 788kS, Greek: 744kS, Estonian: 556kS, Basque: 647kS, Farsi: 738kS, Hindi: 666kS, Croatian: 620kS, Javanese: 622kS, Georgian: 580kS, Khmer: 579kS, Kannada: 434kS, Lithuanian: 554kS, Latvian: 587kS, Malay: 702kS, Mongolian: 629kS, Burmese: 576kS, Min-Nan: 1.2MS, Norwegian: 674kS, Slovak: 618kS, Slovene: 659kS, Serbian: 628kS, Tamil: 507kS, Tagalog: 972kS, Turkish: 627kS & AttractPreserve & Language model \\
\cite{berckmann_low-resource_2020} & UpperSorbian: 600kS & BLEU & Language model, translation \\
\cite{liu_low-resource_2022} & Cantonese: 35kS & SacreBLEU & Translation \\
\cite{pham_van_improving_2022} & Khmer: 70kS & BLEU & Translation \\
\cite{nissanka_exploring_2020} & Tamil: 26kS, Sinhala: 26kS & BLEU & Translation \\
\cite{agarwal_zero-shot_2022} & - & Recall & Question Answering \\
\cite{scalvini_evaluating_2024} & - & Sentence-BERT, BLEU, chrF & - \\
\cite{guo_teaching_2024} & - & COMET, BLEURT, chrF++ & Translation \\
\cite{noauthor_frontiers_nodate} & - & BoolQ, Commitment Bank, COPA, Recognizing Textual Entailment, The Winograd Schema Challenge, classification report & Language model \\
\cite{bendel_llegra_2024} & - & - & - \\
\cite{wu_study_2023} & - & BLEU & Translation \\
\cite{babaali_breaking_2024} & - & - & Translation \\
\cite{li_multi-tasking_2022} & - & BLEU, METEOR, ROUGE, CIDEr & Language model \\
\cite{wang_survey_2022} & - & - & - \\
\cite{mi_multi-granularity_2024} & Uyghur: 20kS & BLEU & Translation \\
\cite{noauthor_efficient_nodate} & - & BLEU & Translation \\
\cite{wang2021survey} & - & - & - \\
\cite{noauthor_english-arabic_nodate} & - & BLEU, chrF, ROUGE & Language model, summarisation, translation \\
\cite{mao_tuning_2024} & Afrikaans: 275kS, Amharic: 89kS, Belarusian: 67kS,  Welsh: 289kS,  Irish: 289kS, Scottish Gaelic: 16kS,  Galician: 515kS, Hausa: 97kS,  Georgian: 377kS, Kazakh: 79kS,  Khmer: 111kS, Kyrgyz: 27kS, Limburgish: 25kS, Burmese: 24kS,  Norwegian Bokmål: 142kS,  Norwegian Nynorsk: 486kS, Occitan: 35kS,  Sinhala: 979kS,  Tajik: 193kS, Turkmen: 13kS,  Tatar: 100kS, Uyghur: 72kS,  Northern Uzbek: 173kS, Eastern Yiddish: 15kS & BLEU, chrF++, COMET & Translation \\
\cite{toraman_llamaturk_2024} & Turkish: 273.9mT & Perplexity, classification report & Language model \\
\cite{zhang_teaching_2024} & Zhuang: 5kS & Bleu, chrF & Translation \\
\cite{acikgoz_bridging_2024} & Turkish: 180GB & ARC-TR, TruthfulQA-TR & Language model, question answering \\
\cite{haque_recent_2021} & - & - & - \\
\cite{shi_low-resource_2022} & - & - & - \\
\cite{zhang_neural_2024} & - & - & - \\
\cite{noauthor_banglagpt_nodate} & Bengali: 26GB & Perplexity & Language model \\
\cite{noauthor_bidirectional_nodate} & - & BLEU, precision, chrF, TER, Rouge & Translation \\
\cite{noauthor_qasina_nodate} & - & Exact match, F1-score, substring match & Question Answering \\
\cite{noauthor_generative_nodate} & - & - & - \\
            \bottomrule
        \end{tabular}
    }
\end{table*}

\end{document}